\documentclass{article}

\usepackage[colorlinks]{hyperref}

\usepackage[accepted]{icml2021}

\usepackage[utf8]{inputenc} 
\usepackage[T1]{fontenc}    
\usepackage{lipsum}
\usepackage[toc,page,titletoc]{appendix}

\usepackage{amsmath,amsthm,amssymb,amsfonts}
\usepackage{microtype}
\usepackage{nicefrac} 
\usepackage{graphicx}
\usepackage{booktabs,multirow}
\usepackage{url}
\usepackage{xcolor}
\hypersetup{
    colorlinks=true,
    linkcolor=blue,
    filecolor=magenta,      
    urlcolor=blue,
    citecolor=blue,
}
\usepackage{wrapfig}
\usepackage{enumitem}
\usepackage{grffile}

\usepackage{subcaption}
\usepackage{stfloats} 

\usepackage{amsmath}
\usepackage{etoolbox}
\usepackage{xcolor}

\usepackage{natbib}   

\usepackage{pifont}

\newcommand{\ft}{\mathbf{f}_{\text{true}}}

\newcommand{\h}{\mathbf{h}}
\newcommand{\q}{\mathbf{q}}
\newcommand{\Dt}{\Delta t}
\newcommand{\dt}{\Delta t}
\renewcommand{\H}{\mathcal{H}}
\newcommand{\bpi}{\boldsymbol{\pi}}

\newcommand{\bphi}{\boldsymbol{\phi}}
\newcommand{\bxi}{\boldsymbol{\xi}}

\newcommand{\z}{\mathbf{z}}

\renewcommand{\v}{\mathbf{v}}

\renewcommand{\a}{\mathbf{a}}
\renewcommand{\t}{\mathbf{t}}
\newcommand{\f}{\mathbf{f}}

\newcommand{\p}{{\mathbf{p}}}

\newcommand{\E}{\mathbb{E}}
\newcommand{\V}{\mathbb{V}}

\newcommand{\R}{\mathbb{R}}
\newcommand{\GP}{\mathcal{GP}}

\newcommand{\N}{\mathcal{N}}

\newcommand{\s}{\mathbf{s}}

\newcommand{\bth}{\boldsymbol{\theta}}

\DeclareMathOperator{\KL}{\textsc{kl}}

\DeclareMathOperator*{\argmin}{arg\,min}

\newcommand{\U}{\mathcal{U}}

\newcommand{\hs}{\hat{\s}}
\newcommand{\ha}{\hat{\a}}
\newcommand{\hf}{\hat{\f}_{\bth}}

\newcommand{\D}{\mathcal{D}}

\icmltitlerunning{Continuous-Time Model-Based Reinforcement Learning}

\begin{document}

\twocolumn[
\icmltitle{Continuous-Time Model-Based Reinforcement Learning}

\begin{icmlauthorlist}
\icmlauthor{\c{C}a\u{g}atay~Y{\i}ld{\i}z}{aalto}
\icmlauthor{Markus~Heinonen}{aalto}
\icmlauthor{Harri L{\"a}hdesm{\"a}ki}{aalto}
\end{icmlauthorlist}

\icmlaffiliation{aalto}{Department of Computer Science, Aalto University, Finland}

\icmlcorrespondingauthor{\c{C}a\u{g}atay~Y{\i}ld{\i}z}{cagatay.yildiz@aalto.fi}

\icmlkeywords{Continuous-time, model-based reinforcement learning, neural ODEs}

\vskip 0.3in
]



\printAffiliationsAndNotice{}  

\begin{abstract} 
Model-based reinforcement learning (MBRL) approaches rely on discrete-time state transition models whereas physical systems and the vast majority of control tasks operate in continuous-time. 
To avoid time-discretization approximation of the underlying process, we propose a continuous-time MBRL framework based on a novel actor-critic method.
Our approach also infers the unknown state evolution differentials with Bayesian neural ordinary differential equations (ODE) to account for epistemic uncertainty. 
We implement and test our method on a new ODE-RL suite that explicitly solves continuous-time control systems.
Our experiments illustrate that the model is robust against irregular and noisy data, is sample-efficient, and can solve control problems which pose challenges to discrete-time MBRL methods.

\end{abstract}

\section{Introduction}

The majority of model-based reinforcement learning (MBRL) methods are based on auto-regressive, discrete-time transition functions that take the current state and action as input and output a new state (or a distribution over it). Since the error incurred by the transition function can be detrimental to learning controls, RL methods have been designed to take \textit{epistemic} uncertainty into account. Two standard ways of doing so have been utilizing ensembles \citep{kurutach2018model,chua2018deep,janner2019trust,pan2020trust} and parametric uncertainty \citep{depeweg2016learning,gal2016improving}.

The most notable deep MBRL breakthroughs have been showcased on discrete-time (DT) problems such as games \cite{hafner2020mastering}, whereas most of the physical and biological systems in science and engineering are inherently continuous in time and follow differential equation dynamics. Despite this fundamental misalignment, discrete-time MBRL has been often applied to continuous-time problems (such as CartPole) by assuming discretized timesteps \citep{brockman2016openai,tassa2018deepmind}.
Indeed, approximating ordinary differential equations (ODE) with the simplest Euler solver where the time increment equals the time difference between measurements matches DT-MBRL and resulting discretized system converges to the true ODE as the discretized time increment approaches zero. Nevertheless,  
accurate approximations come at the cost of increased computational load, a trade-off that is often overlooked. Also, there is no consensus among practitioners on how to choose the correct discretization stepsizes, or how to account for irregular observation times. 

\begin{figure}[t]
    \centering
    \includegraphics[width=\linewidth]{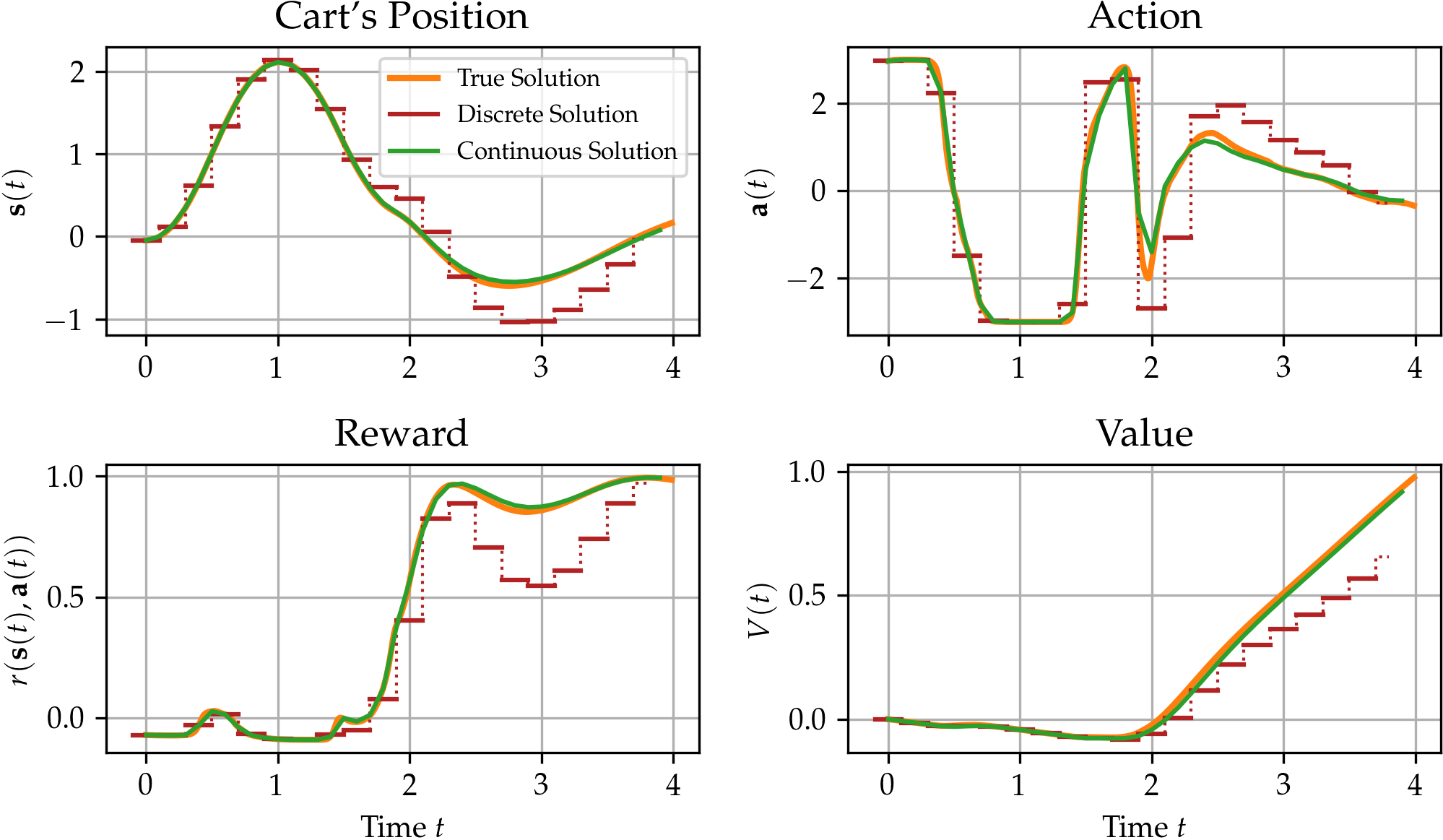}
    \vspace*{-.6cm}
\caption{A comparison of true solution of the CartPole system against discrete and continuous-time trajectories. While continuous-time solution almost perfectly matches the true solution, discrete trajectory diverges from it.}
\label{fig:fig1}
\end{figure}

Aside from discretization issue, adopting continuous-time reinforcement learning setting  \citep{bradtke1994reinforcement,doya2000reinforcement,fremaux2013} presents both conceptual and algorithmic challenges. First, Q-functions are known to vanish in continuous-time systems \citep{baird1994reinforcement,tallec2019making}, which prevents even simple conventional actor-critic or policy learning methods \citep{sutton2018reinforcement} in such a setting.
Second, the auto-regressive one-step ahead state transitions need to be replaced with time derivatives. This has been demonstrated in physics-informed domains such as Lagrangian \citep{lutter2019deep} or Hamiltonian mechanics \citep{zhong2019symplectic}, or in non-RL domains such the pioneering work of neural ODE (NODE) \citep{chen2018neural}. Inference of general-purpose dynamics using ODEs in control settings is still an open question.

In this work, we present a continuous-time model-based reinforcement learning paradigm that carefully addresses all challenges above. Our approach can learn arbitrary time differentials of the governing real-world dynamics, and then forward simulates the surrogate ODE dynamics to learn optimal policy functions. Our key contributions are:
\begin{itemize}
\itemsep0em 
    \item We present a truly continuous-time RL engine, which inherits numerical guarantees of ODE solvers and allows investigation of real-world problems with computer simulations \footnote{Our model implementation can be found at \href{https://github.com/cagatayyildiz/oderl}{https://github.com/cagatayyildiz/oderl}}.
    \item We describe novel strategies to handle epistemic uncertainty in neural ODE models and successfully infer black-box controlled ODEs.
    \item We present a novel theoretically consistent CT actor-critic algorithm that generalizes existing state-value based policy learning methods.
    \item We demonstrate our method on three different RL problems and present its robustness to both noisy and irregular data, which poses major challenges to conventional discrete-time methods.
\end{itemize}

\section{Continuous-Time Reinforcement Learning}

We consider continuous-time (CT) control systems governed by an ordinary differential equation (ODE)
\begin{equation}
    \dot{\s}(t) = \frac{d \s(t)}{dt} = \f\big(\s(t),\a(t)\big), \label{eq:ode-def}
\end{equation}
where $\s(t)\in \R^d$ denotes a vector-valued \textit{state} function at time $t$, and $\a(t)\in \R^m$ is the \textit{control} input function, or action. The function $\f : \R^{d+m} \mapsto \R^{d}$ maps the current state-action pair $(\s,\a)$ to a state derivative $\dot{\s}$.

The state at real-valued time $t \in \R_+$ depends on the initial state $\s_0$ as well as the infinitesimal sequence of past actions $\a[0,t)$, and can be solved with
\begin{equation*}
    \s(t) = \s_0 + \int_0^t \f\big(\s(\tau),\a(\tau)\big) d\tau, \label{eq:ode-sol}
\end{equation*}
where $\tau \in \R_+$ is an auxiliary time variable. 

The main objective of CT control is to maximize the reward integral
$ \int_0^\infty r\big( \s(\tau), \a(\tau) \big) d\tau, $
which we translate instead into maximizing the discounted infinite horizon reward integral \citep{sutton2018reinforcement}, or \emph{value} function, 
\begin{align}
    V(\s(t)) &= \int_t^\infty e^{-\frac{\tau-t}{\eta}}  r\big(\s(\tau),\a(\tau)\big) d\tau, \label{eq:value}
\end{align}
where $r:\R^{d+m}\rightarrow\R$ is an instantaneous reward and $0<\eta<1$ is the discount factor. The optimal control sequence $\a^*[t,\infty]$ maximizing $V(\s(t))$ is usually bounded by $[\a_{\text{min}}, \a_{\max}]$.

The control problem \eqref{eq:value} is a constrained function optimisation problem, which is generally intractable due to system non-linearity and continuous dynamics \citep{bertsekas1995dynamic}. For very limited cases such as linear state feedback $\a=K\s+\mathbf{v}$, the solution can be analytically obtained by Pontryagin’s maximum principle \citep{athans2013optimal} or dynamic programming \citep{bellman1966dynamic,bertsekas1995dynamic}. For an overview of numerical approaches to solve continuous-time optimal control problems, see e.g.~\cite{diehl2017numerical}.

In this paper, we approach the control problem from a reinforcement learning standpoint. In particular, our main goal is to find a \emph{policy} function
\begin{align}
\a(t)=\bpi(\s(t)) \label{eq:policy}
\end{align}
that maximizes the values $V(\s(t))$. This problem formulation signifies a fundamental separation from standard RL methods in several ways.
\begin{itemize}
\itemsep0em 
\item First, for the dynamics differential $\dot{\s}(t)$ \eqref{eq:ode-def} to exist, the function $\f$ needs to be differentiable over time. Since the existing MBRL methods typically approximate the unknown true CT dynamics via DT stochastic function approximators such as Gaussian processes \citep{deisenroth2011pilco}, Bayesian neural networks \citep{gal2016improving,depeweg2016learning} or probabilistic neural networks \cite{chua2018deep}, they cannot be trivially converted into continuous-time.
\item Second, deterministic dynamics is also difficult to match with non-deterministic policies \eqref{eq:policy}. Therefore, standard policy gradient methods that require stochastic policies, such as TRPO \citep{schulman2015trust} and PPO \citep{schulman2017proximal}, are no longer applicable. \item Finally, since the Q-function is ill-defined in continuous time \cite{baird1994reinforcement}, Q-learning based policy methods, such as DDPG \citep{lillicrap2015continuous}, SAC \citep{haarnoja2018soft} and MBPO \cite{janner2019trust}, are eliminated.
\end{itemize}

\subsection{Earlier works}
The earliest work in CTRL literature, named advantage updating, was pioneered by \citet{baird1993advantage} and later extended by \citet{tallec2019making}. This model-free algorithm learns the advantage and state-value functions by discretizing the value integral $V(\cdot)$ \eqref{eq:value}. The optimal policy is then given by the action that maximizes the advantage function for any input state. In control literature continuous-time systems have been studied under limited linear-quadratic (LQ) control setting \citep{wang2020,modares2014}, while \citet{ghavamzadeh2001continuous} proposed a CT hierarchical RL method for distributed value functions. Recent work also explores optimal control of physics-informed CT neural dynamical systems such as Lagrangian \citep{lutter2019deep} and Hamiltonian mechanics \citep{zhong2019symplectic}. Finally, \citet{du2020model} present a CT-MBRL framework for semi-Markov decision processes (MDPs) with a latent ODE-RNN dynamics model and a Q-learning approach adapted for semi-MDPs. A through overview of related work is given in supplementary Section~S4.

The closest to our work is the seminal paper by \citet{doya2000reinforcement}, which describes a technique to learn optimal policy in continuous-time while respecting above-mentioned limitations. The method approximates the state-value function with a radial basis function and Euler discretization of the value integral \eqref{eq:value}. An actor-critic algorithm based on temporal difference error was also proposed. \citet{fremaux2013} adapt the model to spiking neural networks. 

Our approach, outlined in Algorithm~\ref{alg:ctrl}, differs fundamentally from these approaches in two respects: \emph{(i)} we estimate the unknown dynamics using non-linear neural ODEs, and \emph{(ii)} explicitly forward integrate the dynamics for non-linear policy learning. Next we detail the components of our model, namely, the dynamics model and continuous-time actor-critic algorithm.


\begin{figure*}[t]
    \centering
    \begin{subfigure}[b]{.48\textwidth}
        \centering
        \includegraphics[width=\textwidth]{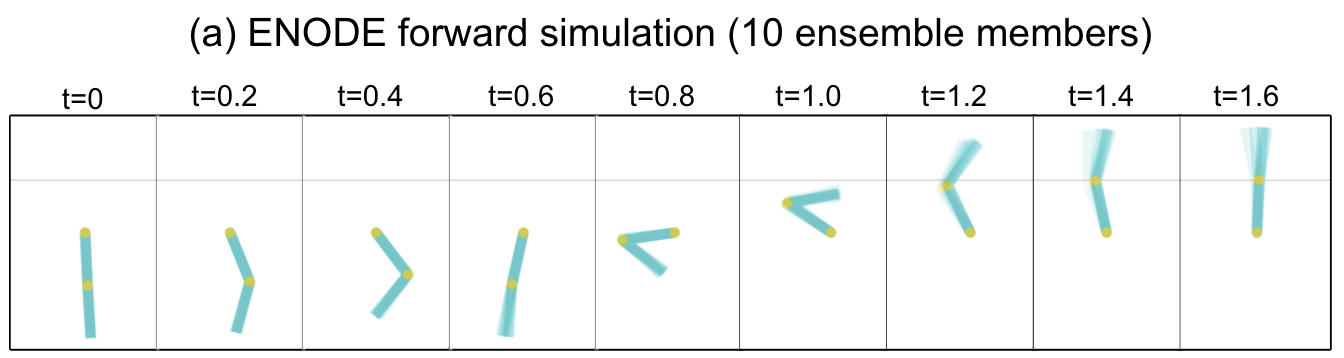}
        \phantomsubcaption
        \label{fig:acrobot1}
    \end{subfigure}
    \hfill
    \begin{subfigure}[b]{.48\textwidth}
        \centering
        \includegraphics[width=\textwidth]{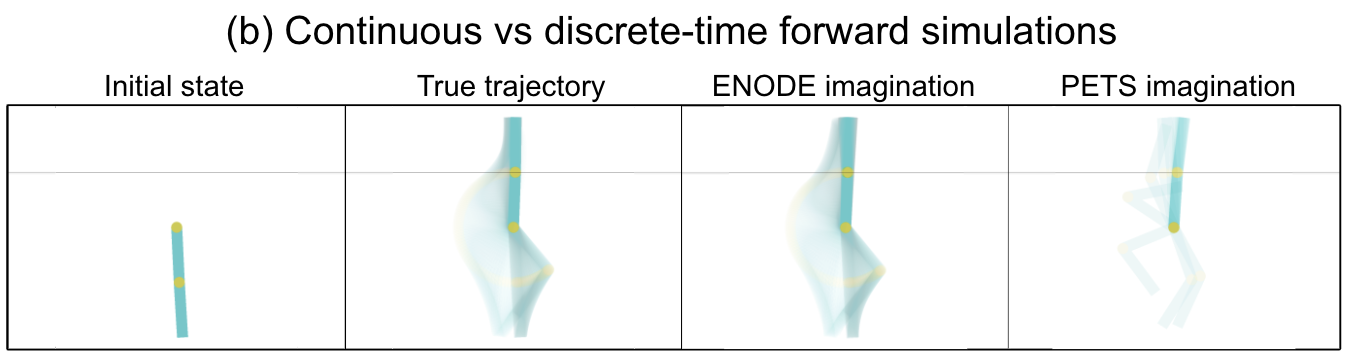}
        \phantomsubcaption
        \label{fig:acrobot2}
    \end{subfigure}
    \vspace*{-0.6cm}
    \caption{An illustration of our ENODE dynamics model on Acrobot task. \textbf{(a)} Ten trajectory samples computed by \eqref{eq:dyn-int} at uniform intervals. \textbf{(b)} The continuous path from initial state to goal state using the true and ENODE dynamics as well as PETS' discrete transition model (single ensemble member is plotted).}
\end{figure*}

\section{Dynamics Learning}
\label{sec:dyn}

We start by assuming that the dataset consists of $N$ observed state-action trajectories $\D = \{(\s_{0:T}^{(n)},\a_{0:T}^{(n)})\}_{n=1}^N$ collected from a CT environment we aim to model or, in case of simulated environments, obtained by solving the true ODE system at observation time points $t_{0:T}$. We denote the $i$'th observation within a trajectory by $\s_i:=\s({t_i})$, and $\dt_i=t_{i+1}-t_i$ stands for the time until the next observation $\s_{i+1}$. We note that observations may arrive irregularly in time.

\subsection{Previous Work}

The dynamics models in state-of-the-art MBRL techniques such as PILCO \citep{deisenroth2011pilco} and PETS \citep{chua2018deep} approximate the state difference $\Delta \s_{t} = \f(\s_t,\a_t)$ such that $\s_{t+1} = \s_t + \Delta \s_t$. Such approximations are implicitly based on uniform sampling assumption. To see that, imagine the system visits some state twice $\s_i=\s_j=\s$ at two timepoints $t_i$ and $t_j$, and the same action $\a$ is applied. If the observations arrive non-uniformly, i.e., $\dt_i \neq \dt_{j}$, then the next states would be different: $\s_{i+1}\neq\s_{j+1}$ (given the trivial condition $||\f(\s)|| \neq 0$). Consequently, resulting data triplets for dynamics learning would have the same inputs $(\s,\a)$ and different targets, which would lead to training issues. 

To adapt these discrete-time methods for temporally irregular trajectories, we give the time increment $\Delta t_i$ as an additional parameter to the dynamics transition,
\begin{equation}
    \s_{i+1} = \s_i + \f(\s_i,\a_i, \Delta t_i). \label{eq:mod-pets}
\end{equation}
We refer this version with \emph{modified} prefix and use \emph{vanilla} for the traditional way of training. 

\begin{algorithm*}[t]
   \caption{Continuous-Time MBRL with Neural ODEs}
   \label{alg:ctrl}
\begin{algorithmic}[1]
    \STATE Choose a variational approximation for NODE model and initialize dynamics $\f_{\bth}$, actor $\bpi_{\bphi}$ and critic $\nu_{\bxi}$ estimates
    \STATE Collect a dataset $\D = \{(\s_{0:T}^{(n)},\a_{0:T}^{(n)})\}_{n=1}^{N_0}$ of $N_0$ trajectories with smooth random policies $\pi^{(n)}(\s,t)\sim\GP(\mathbf{0},\mathbf{k}(t,t'))$
    \REPEAT
    \STATE \texttt{// Dynamics learning}
    \FOR{$i=1...N_\text{dyn}$}
        \STATE Randomly sample from $\D$ a mini-batch of $N_\text{d}$ subsequences, each of which of length $t_\text{s}=5$: \\ $\D' = \{(\s_{t_n:t_n+t_\text{s}}^{(n)},\a_{t_n:t_n+t_s}^{(n)})\}_{n \in \Xi}, ~~ \Xi \subset \{1,\ldots,N\}, ~~ |\Xi|=N_\text{d}, ~~ t_n \sim \U[0,T-t_s]$
        \STATE Update $\bth$ by taking a gradient step using the \textsc{ELBO} on $\D'$ by (\ref{eq:elbo},\ref{eq:mc},\ref{eq:dyn-int})
    \ENDFOR
    \STATE \texttt{// Actor-Critic learning}
    \FOR{$i=1...N_\text{ac}$}
        \STATE Uniformly sample $N_p=100$ states $\{\s^{(p)}\}_{p=1}^{N_p}$ from the most recently collected ten data sequences
        \STATE Imagine trajectories $\hat{\s}$ from each $\s^{(p)}$ using current estimates of dynamics $\hat{\f}_{\bth}$ and policy $\bpi$ by (\ref{eq:img-st},\ref{eq:img-act}).
        \STATE Update $\bphi$ and $\bxi$ by taking gradient steps using the imagined trajectories by (\ref{eq:actor-loss},\ref{eq:critic-loss})
    \ENDFOR
    \STATE Collect data $\D \leftarrow \D \cup \{(\s_t,\a_t)_{t=0}^{T}\}$ by interacting with real world: $\s_{0:T},\a_{0:T} \leftarrow \s_0 + \int_0^T \ft(\s(\tau),\bpi(\s(\tau)))d\tau$  
    \STATE [optional] Collect more sequences with an exploring policy: $\bpi^{\text{exp}}\left(\s(t),t\right) = \bpi\left(\s(t)\right)+\z(t), ~ \z(t) \sim \mathcal{GP}(\boldsymbol{0},\mathbf{k}(t,t'))$ 
    \STATE Execute the policy in the real environment with ten different initial configurations
    \UNTIL{The pole never falls once upright in ten trials}
\end{algorithmic}
\end{algorithm*}

\subsection{CT Dynamics with Bayesian Neural ODEs}
We consider deep neural ODE models $\hat{\f}_{\bth}$ \citep{chen2018neural}, where $\bth$ denotes the network parameters. Our goal is to estimate the state differential $\hat{\f}_{\bth}$ such that its forward-simulated ODE trajectories reproduce the true ODE trajectories accurately. 
Our main learning objective is to compute the posterior distribution $p(\bth | \D)$. Unfortunately, the exact posterior in black-box ODE models are intractable and typically MAP estimates are computed \citep{heinonen2018learning,chen2018neural}. Since handling epistemic uncertainty to reduce model bias lies at the heart of MBRL \citep{deisenroth2011pilco,chua2018deep}, we resort to variational inference by introducing an approximate posterior distribution $q(\bth)$ closest to the true posterior in KL sense,
\begin{align*}
    \argmin_{q(\bth)} \KL[ q(\bth) \, || \,  p(\bth | \D)].
\end{align*}
Finding the optimal variational approximation is equivalent to  maximizing the evidence lower bound (ELBO) \citep{blei2017variational}:
\begin{align}
    \log p(\mathcal{D}) &\geq \E_q\Big[\log p(\mathcal{D} | \bth)\Big] - \KL\big[q(\bth) \, || \, p(\bth)\big], \label{eq:elbo}
\end{align}
where $p(\bth)$ is the parameter prior. Note that the likelihood $p(\mathcal{D}|\bth)$ is evaluated only on the state-trajectories, i.e., $p(\mathcal{D}|\bth) = p(\mathcal{D}_{\s} | \mathcal{D}_{\a},\bth)$, where $\mathcal{D_{\s}} = \{\s_{0:T}^{(n)}\}_{n=1}^N$ and $\mathcal{D_{\a}} = \{\a_{0:T}^{(n)}\}_{n=1}^N$. An unbiased estimate of the expected log-likelihood can be obtained by Monte Carlo integration,
\begin{align}
    \E_q\big[\log p(\mathcal{D}|\bth)\big] \approx \frac{1}{L} \sum_{l=1}^L \sum_{n=1}^N \sum_{i=0}^T \log  \N\left(\s_i^{(n)} | \hs_i^{(n,l)},\Sigma\right), \label{eq:mc}
\end{align}
where $\Sigma$ denotes trainable observation noise variances. A trajectory sample $\hs_{0:T}^{(n,l)}$ is drawn by first sampling from the posterior $\bth^{(l)} \sim q(\bth)$ and then forward simulating the dynamics model $\hat{\f}_{\bth^{(l)}}(\s,\a)$:
\begin{align}
    \hs_i^{(n,l)}\mid\big(\s_0^{(n)},\a_{0:t}^{(n)}\big) = \s_0^{(n)} + \int_0^{t_i} \hat{\f}_{\bth^{(l)}} \big(\hs^{(n,l)}_\tau,\a^{(n)}_\tau \big) d\tau. \label{eq:dyn-int}
\end{align}
We demonstrate trajectory samples in Figure~\ref{fig:acrobot1}.

\paragraph{Time-continuous actions} The state trajectory in \eqref{eq:dyn-int} is fully determined by the initial state $\s_0$ and the infinitesimal actions $\a[0,t)$ applied during integration. Since the integral may be evaluated at arbitrary time points, the actions must be continuous in time, whereas they are assumed to be known only at observation time points. A simple yet convenient mean to obtain time-continuous actions is interpolating between observed discrete actions. We opt for kernel interpolation with a squared exponential kernel function whose length-scale parameter can be set depending on the smoothness of the action sequence. GP-based stochastic interpolations would also be considered to reduce the bias incurred by interpolation.

\paragraph{Variational approximation} If the variational posterior is chosen to be a mixture of Dirac densities on the weights, the resulting approximation corresponds to an ensemble of neural ODEs (ENODE). Similar to other ensemble methods, each member would learn a different vector field, which provide means for epistemic uncertainty modeling. Alternatively, $q(\bth)$ can be defined on the weights, hidden neurons or function outputs, which would lead to weight-space \citep{yildiz2019ode2vae}, implicit \citep{trinh2020scalable} or functional \citep{sun2019functional} Bayesian neural ODE models.


\section{Continuous-Time Actor-Critic Algorithm}
In this section, we describe an actor-critic algorithm that is fully compatible with the CT framework. Our algorithm relies solely on the forward simulation of the inferred dynamics, i.e., no interaction with the real-world is required, making it much easier to deploy to real-world compared to the standard MB controllers such as MPC \citep{chua2018deep} and policy learning methods such as MBPO \citep{janner2019trust}. Also, the standard discrete-time temporal difference (TD) learning methods \citep{schulman2015trust,sutton2018reinforcement,hafner2019dream} appear as special cases of our CT actor-critic method as shown later.

Our main policy learning goal is to maximize the discounted cumulative reward when the policy is executed in the real-world, which simply amounts to computing the integral in \eqref{eq:value}. Because the true differential function is unknown, we leverage our proxy dynamics $\hf$ to maximize a surrogate objective:

\begin{align}
    V(\s_0) \approx \hat{V}(\s_0) = \E_{q(\bth)} \left[ \int_0^\infty e^{-\frac{\tau}{\eta}} r(\hs(\tau),\ha(\tau))d\tau \right] \label{eq:value-sur}
\end{align} 
where $\hs[0,t)$ and $\ha[0,t)$ are referred to as \textit{imagined trajectory} and \textit{imagined actions}\footnote{The actions are applied to imagined states, rendering the actions imagined as well.} and $\s_0$ denotes an initial value. 
The imagination is controlled by a deterministic policy function $\bpi_{\bphi}(\cdot)$ with parameters $\bphi$:
\begin{align} 
    \hs(t) &= \s_0 + \int_0^t \hf\big(\hs(\tau), \ha(\tau)\big) \label{eq:img-st} d\tau  \\
    \ha(t) &= \bpi_{\bphi}(\hs(t)). \label{eq:img-act}
\end{align}

Note that this formulation comes with the differentiable reward assumption. The reward function typically consists of a sum $r(\s,\a)=d(\s,\s^*)+c||\a||_2$, where the first term measures some distance between the current state and a target state $\s^*$, and the latter term penalizes the action magnitude. In turn, policy learning task can be informally stated as finding a set of policy parameters $\bphi$ that drives the system towards high-reward states while applying smallest possible actions. If the initial state $\s_0$ follows a distribution $\s_0\sim p(\s_0)$, then the optimization target becomes $\E_{p(\s_0)}[\hat{V}(\s_0)]$, which is usually approximated by Monte-Carlo averaging.

\paragraph{Critic}
The optimization objective in \eqref{eq:value-sur} requires computing an infinite integral. To circumvent this problem, we first re-write the value function in a recursive manner: 
\begin{align*}
    \hat{V}^H(\s_0) &= \E_{q(\bth)} \left[\int_0^H e^{-\frac{\tau}{\eta}} r(\hs(\tau),\ha(\tau))d\tau + e^{-\frac{H}{\eta}}\hat{V}(\hs_H) \right]
\end{align*}
where $H$ is known as \textit{horizon}, $\hs_H := \hs(H)$ is the end-state of the integral, and $\hat{V}(\hs_H)$ is \textit{future reward}. The first term on the rhs represents the imagined reward, which can be computed by evaluating the reward function on the trajectories given by (\ref{eq:img-st}-\ref{eq:img-act}), whereas the second term is an infinite integral. Next we introduce the critic, or state-value approximation, that replaces intractable future reward term:
\begin{align}
    \hat{V}(\s) \approx \nu_{\bxi}(\s), \label{eq:critic}
\end{align} 
where $\nu_{\bxi}$ is a neural network with parameters $\bxi$. 
With a slight abuse of notation, we rewrite the value function as follows:
\begin{align}
    \hat{V}^H(\s_0) &= \E_{q(\bth)} \left[\int_0^H e^{-\frac{\tau}{\eta}} r(\hs(\tau),\ha(\tau))d\tau + e^{-\frac{H}{\eta}}\nu_{\bxi}(\hs_H) \right]. \label{eq:value2} 
\end{align}
Consequently, our new objective seeks to learn a policy function $\bpi_{\bphi}$ that maximizes the reward integral over horizon $H$ and ends up at a high-value state. 

\paragraph{Learning the actor} The actor aims to maximize the value estimates \eqref{eq:value2} for a batch of initial values. 
In order to learn the optimal policy everywhere, initial values are drawn from the previous experience. More formally, we solve the following maximization problem:
\begin{align}
    \max_{\bphi} ~ \E_{q(\s)} \left[ \hat{V}^H(\s) \right], \quad q(\s) = \frac{1}{|\D|} \sum_{n,i} \delta\left(\s-\s_i^{(n)}\right). \label{eq:actor-loss}
\end{align}
Observe that both the reward integral and the future reward in \eqref{eq:value2} depend on the imagined state trajectory, which in turn depends on the actions. Therefore, the gradient of \eqref{eq:actor-loss} w.r.t.\ $\bphi$ can easily be computed in auto-differentiation frameworks that support ODE gradients.

\paragraph{Learning the critic} Similar to discrete-time frameworks, our method utilizes temporal difference learning to estimate the state-value function in \eqref{eq:critic}. In particular, the value approximation $\nu_{\bxi}(\cdot)$ is regressed on the sum of the finite reward integral and future reward given in \eqref{eq:value2}. Since the value approximation holds for any horizon length $H$, we propose to marginalize it out:
\begin{align}
    \min_{\bxi} ~ \E_{q(\s)} \left[ \left( \nu_{\bxi}(\s) - \E_{q(h)} \left[ \hat{V}^h(\s) \right] \right)^2 \right], ~~ q(h) = \U[0,H] \label{eq:critic-loss}
\end{align}
where the outer expectation is w.r.t.\ $q(\s)$ shown in~\eqref{eq:actor-loss},  $\U$ denotes the uniform distribution, and the inner expectation is approximated by Monte-Carlo sampling. We use a separate target network that generates the future rewards, an idea that is known to stabilize the critic training \citep{mnih2015human}. We update the target network with a copy of the critic every 100 optimization iterations. 

Both actor and critic loss functions require trajectory sampling from the dynamics model. In practice, we forward integrate the dynamics model once and utilize the resulting state-action trajectories to optimize both objectives. The inner expectation in \eqref{eq:critic-loss} can be evaluated using the intermediate states without incurring any additional overhead.

\paragraph{Connection to temporal difference} Discrete state-value formulation can be retrieved by discretizing the reward integral \eqref{eq:value} with a predetermined step size 
$
    \overline{V}(\s_t) = \sum_{l=0}^\infty \gamma^l r(\s_{t+l},\a_{t+l})
$
where the overbar sign denotes discrete formulation and $\gamma$ is the discount factor. In such a scenario, $\hat{V}^h(\s)$ would reduce to so-called $h$-step return: $\overline{V}^h(\s_t)=\sum_{l=0}^{h} \gamma^l r_{t+l} + \overline{V}(\s_{t+h})$. Marginalizing the horizon would correspond to computing the mean of the $h$-step returns, a commonly used technique to reduce the gradient variance \citep{schulman2015high}. Finally, TD-$h$ learning would be retrieved by simply keeping the horizon fixed at $h$, i.e., with $q(h)$ being a Dirac distribution. As demonstrated in Figure~\ref{fig:acrobot2}, DT models compute the value estimates using a finite collection of states $\s_{t:t+h}$, and thus are inaccurate.

\begin{figure}[t]
    \centering
    \includegraphics[width=\linewidth]{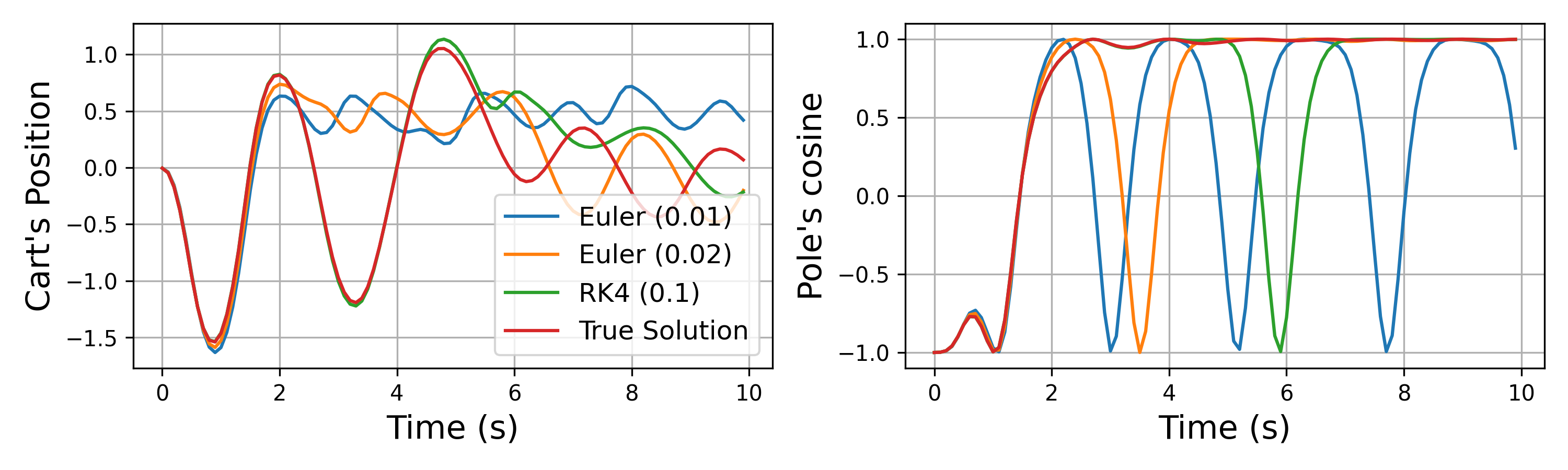}
    \vspace*{-0.8cm}
    \caption{Comparison of different discretization choices from \cite{brockman2016openai, gal2016improving, tassa2018deepmind} on CartPole swing-up task vs.\ the true ODE solution. Discretization timesteps are given inside the parenthesis.}
    \label{fig:discretization-comp}
\end{figure}

\section{Experiments}
We experimentally evaluate our model on three CTRL benchmark environments: Pendulum, CartPole and Acrobot. In all tasks, the initial state is ``hanging down'' and the goal is to swing up and stabilise the pole at the target position where the pole stays in an upright position (also at the origin in CartPole task). We define the differentiable reward functions as the exponential of the distance to target state, while also penalizing the magnitude of the velocity and action. The actions are assumed to be continuous and bounded. The experiments aim to answer the following questions:
\vspace*{-.3cm}

\begin{itemize}
\itemsep0em 
    \item Do existing RL frameworks simulate CT dynamics?
    \item Do DTRL methods fail with irregularly sampled data?
    \item Is our ENODE approximation sensitive to the imagination horizon and the number of ensemble members?
    \item How does our model perform in standard RL tasks?
    \item How does the model performance change with \textit{(i)} noisy data, \textit{(ii)} irregular sampling, and \textit{(iii)} various $\Dt$?
\end{itemize}

\paragraph{Compared models} We consider three different variational approximations for our model: Ensemble neural ODE (ENODE), batch ensemble neural ODE (BENODE) and implicit Bayesian neural ODE (IBNODE). ENODE maintains a separate MLP for each ensemble member \citep{lakshminarayanan2016simple} while BENODE learns shared weight parameters as well as member-specific rank-1 multipliers \citep{wen2020batchensemble}. IBNODE considers a mixture of Gaussians variational posterior on layer-wise multiplicative factors \citep{trinh2020scalable}. In addition, we include two DT transition models, namely, modified PETS and deep PILCO \citep{gal2016improving}. Policy learning using DT transition models is performed by discretizing the integrals in (\ref{eq:actor-loss},\ref{eq:critic-loss}).

\paragraph{Observation spacing} In standard RL frameworks such as OpenAI Gym, the observations arrive at constant intervals $\Dt=\kappa$ \citep{brockman2016openai}. In addition to this baseline, we consider two irregular sampling scenarios in which observations \textit{(i)} arrive uniformly within a range $\Dt \sim \U(0, 2\kappa]$, or \textit{(ii)} follow an exponential distribution $\text{Exp}(\kappa)$ with $\kappa$ being the scale parameter. Note that all distributions have the same mean time increments $\kappa$.

\paragraph{Additional details} We refer to each iteration of the while loop in Algorithm~\ref{alg:ctrl} as a \emph{round}. Following \citet{deisenroth2011pilco}, the model performance is measured at the end of each round by executing the inferred policy in the real world for $H=30$ seconds. Our main result Figures~\ref{fig:comparison5}-\ref{fig:ablations}-\ref{fig:comparison2} illustrate mean rewards and 90/10 percent quantiles, computed over 20 experiments. Each method is given the same amount of computational budget; therefore, the number of completed rounds differ among methods. We implement our method in PyTorch \citep{paszke2019pytorch} and use adaptive checkpoint adjoint method to backpropagate through the ODE solver \citep{zhuang2020adaptive}. Please refer to supplementary Section~S2 for more details.

\begin{figure}[t]
    \centering
    \includegraphics[width=\linewidth]{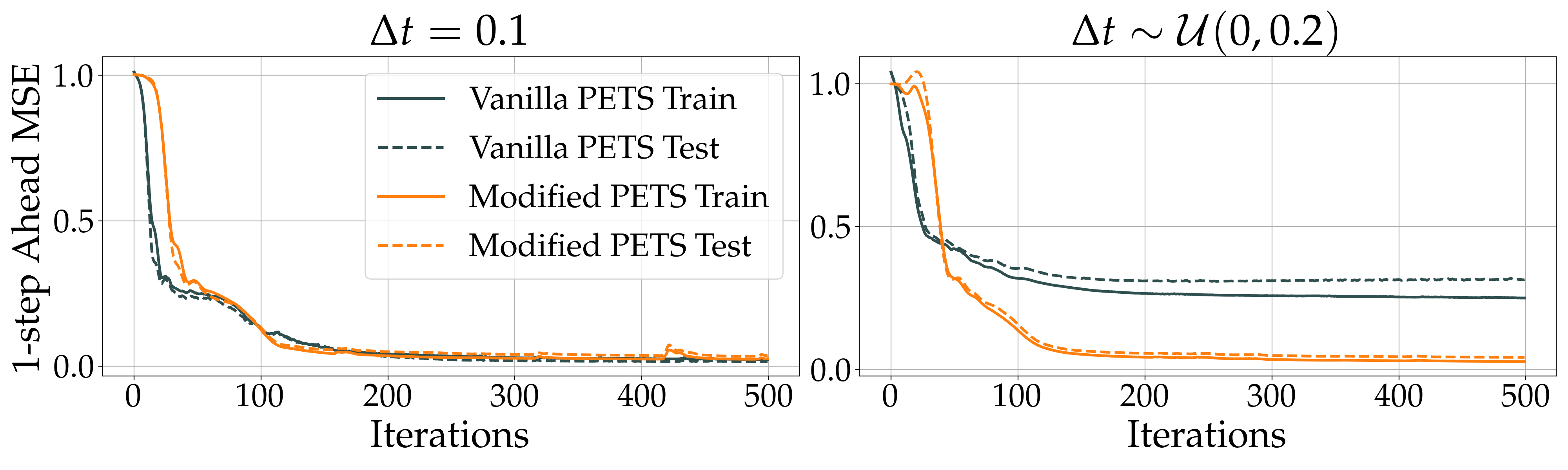}
    \vspace*{-0.8cm}
    \caption{Comparison of the vanilla and modified versions of PETS dynamics on regularly and irregularly sampled data. Vanilla PETS cannot handle irregularly sampled data while model performances match on the former case.}
    \label{fig:pets-comp}
\end{figure}

\subsection{A New CTRL Framework}
The test environments are already implemented in standard RL frameworks such as OpenAI Gym \citep{brockman2016openai} and DeepMind Control Suite \citep{tassa2018deepmind}. Existing implementations either perform discrete transitions or inaccurate numerical integration routines, e.g., fixed-step ODE solvers with too big step sizes. Furthermore, as illustrated in Figure~\ref{fig:fig1}, the actions in our CT framework smoothly change over time whereas DT approximations implicitly assume piece-wise constant control signals. Figure~\ref{fig:discretization-comp} illustrates how different discretization choices result in distinct state trajectories on CartPole problem: While the policy is able to swing-up and balance the pole in the upright position when integrated with the correct solver, crude Euler and RK4 approximations bifurcate near the position where the pole hangs up. 

To deal with above-mentioned inaccuracies, we implement a new CTRL framework which takes as input a deterministic policy function $\bpi(\cdot)$ and observation time points, performs forward integration in a numerically-stable way, and returns the state trajectory at corresponding time points. 
Empirical evidence for the need of integration as well as a comparison of different ODE solvers are presented in supplementary Section~S3.

\begin{figure*}[t]
\hfill
\begin{minipage}[t]{0.33\textwidth}
    \centering
    \textsc{Pendulum}
    \includegraphics[width=\linewidth]{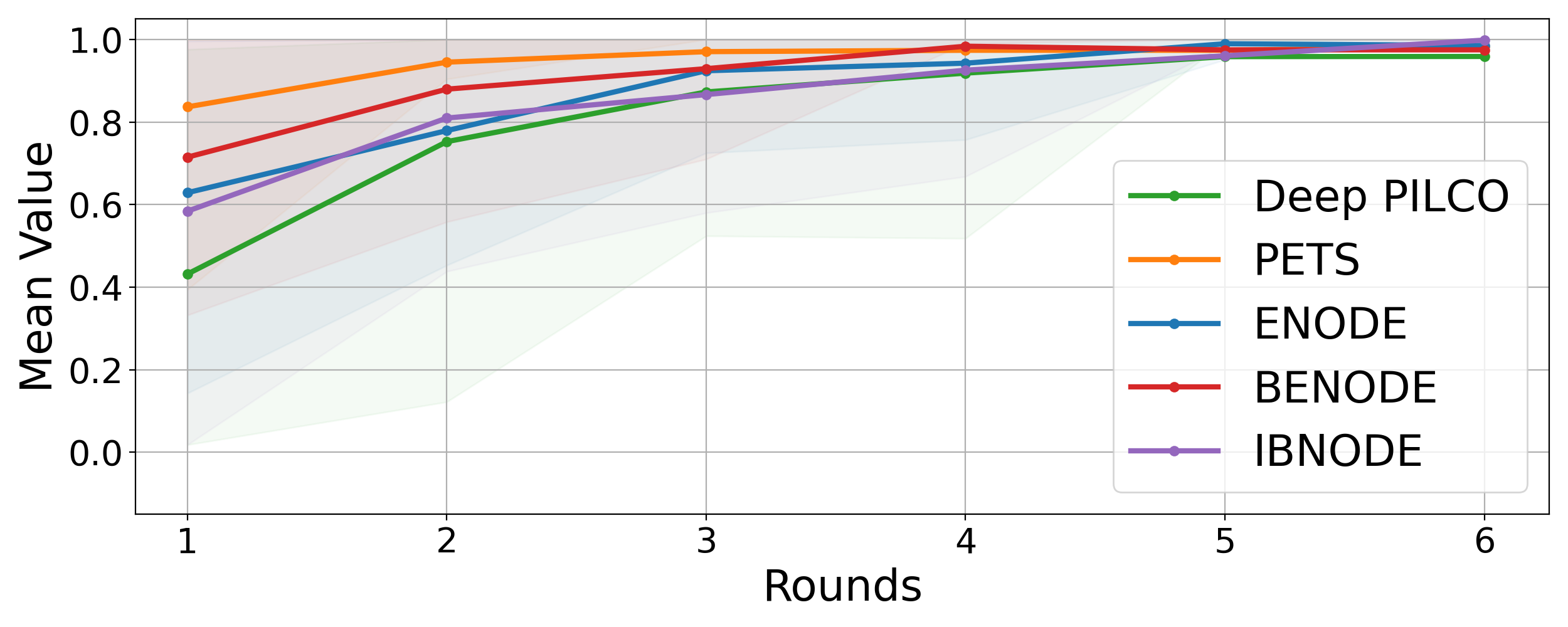}
\end{minipage}
\begin{minipage}[t]{0.33\textwidth}
    \centering
    \textsc{CartPole}
    \includegraphics[width=\linewidth]{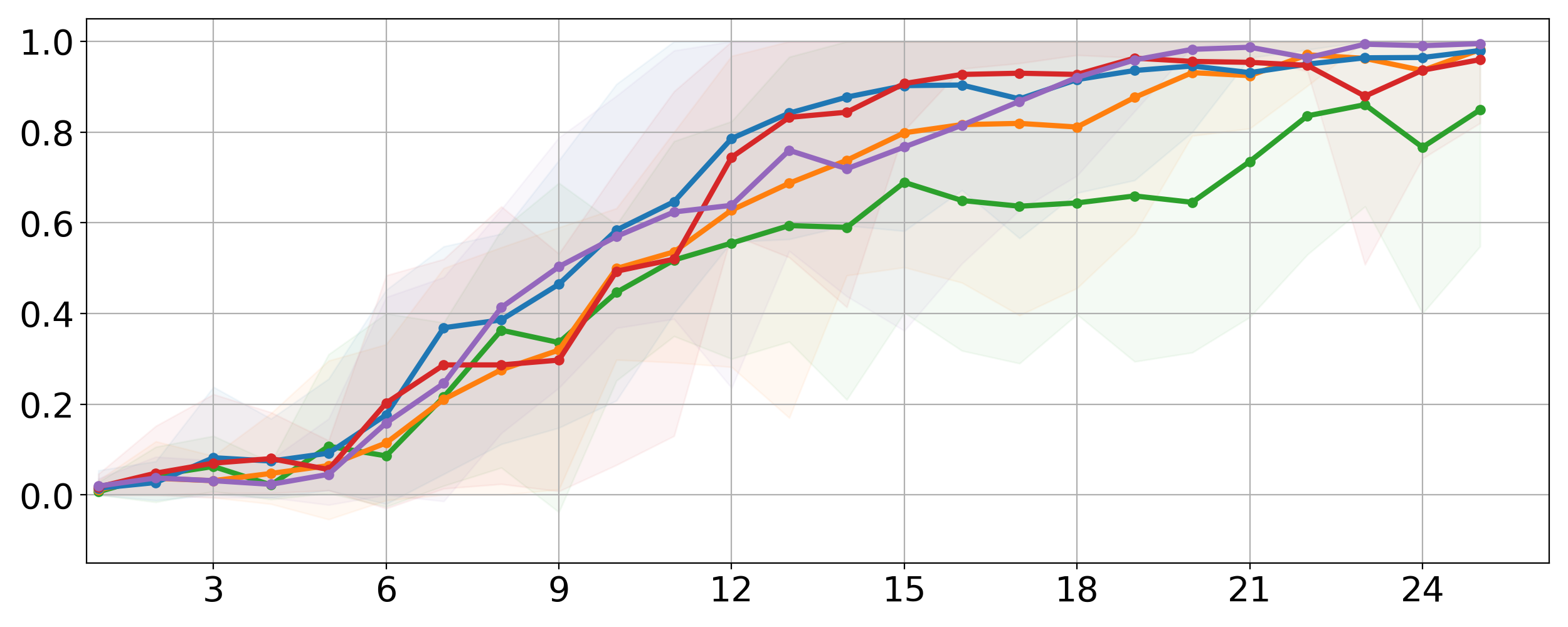}
\end{minipage}
\begin{minipage}[t]{0.33\textwidth}
    \centering
    \textsc{Acrobot}
    \includegraphics[width=\linewidth]{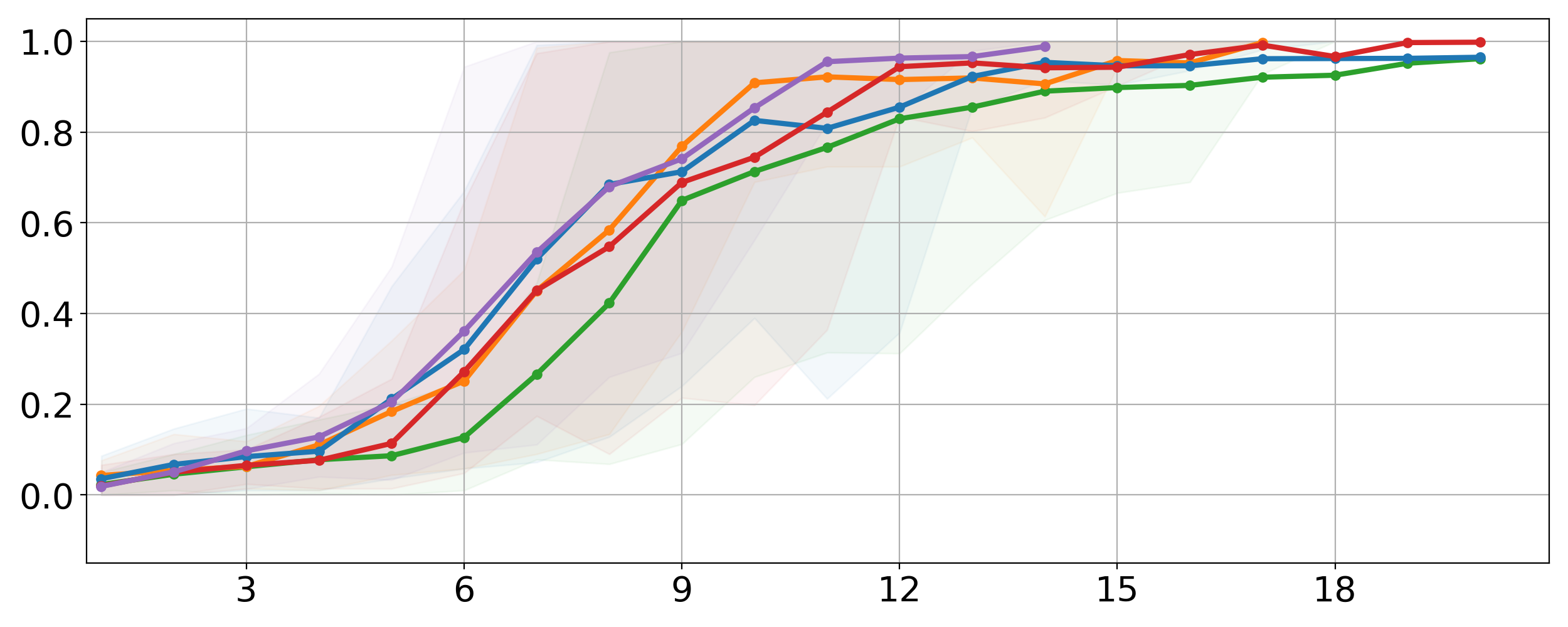}
\end{minipage}
\hfill
\begin{minipage}[t]{0.33\textwidth}
    \centering
    \includegraphics[width=\linewidth]{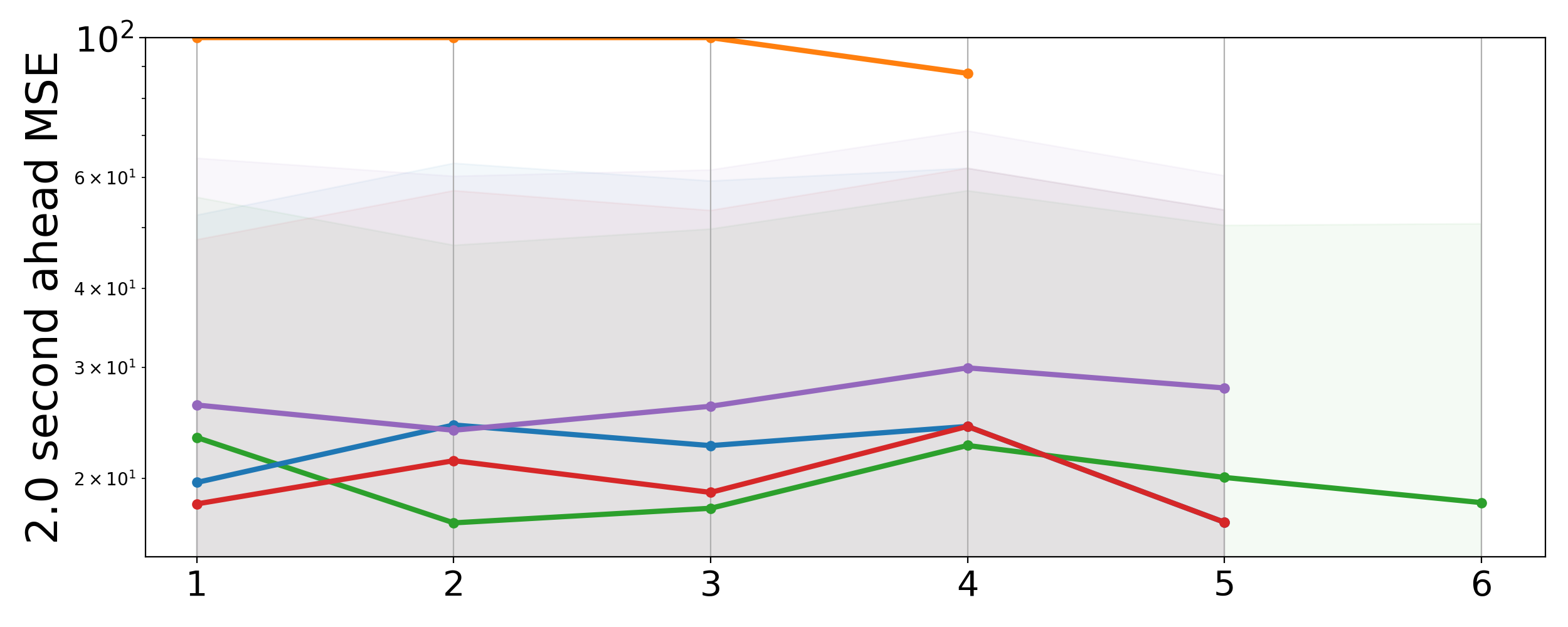}
\end{minipage}
\begin{minipage}[t]{0.33\textwidth}
    \centering
    \includegraphics[width=\linewidth]{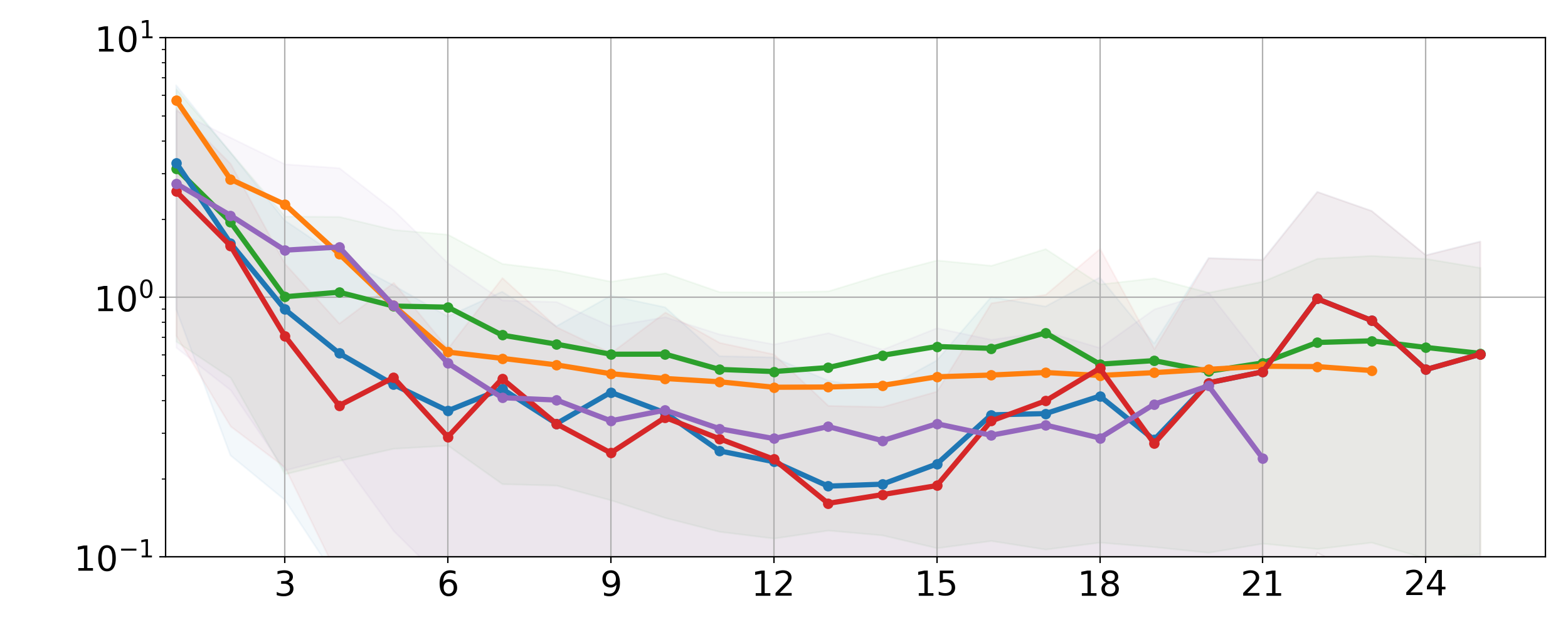}
\end{minipage}
\begin{minipage}[t]{0.33\textwidth}
    \centering
    \includegraphics[width=\linewidth]{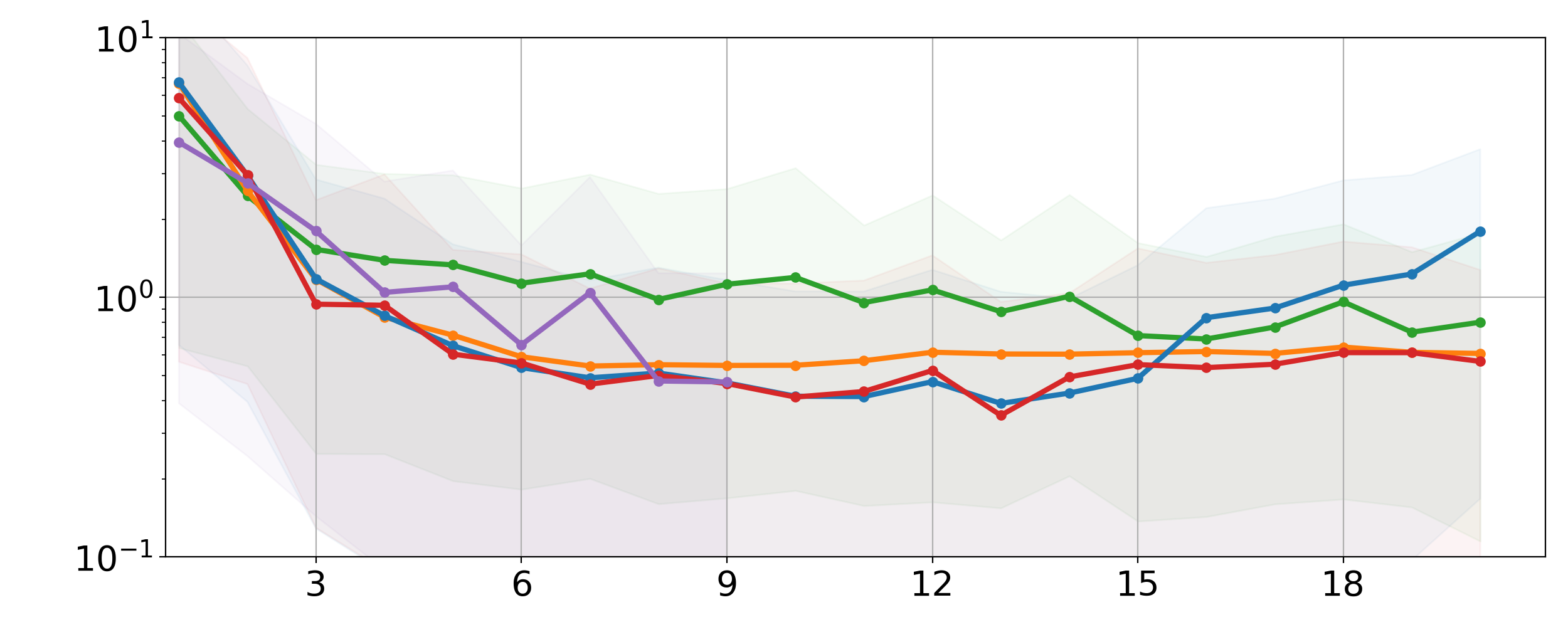}
\end{minipage}
\caption{A comparison of three variants of our model and two other dynamics approximations \citep{gal2016improving,chua2018deep} on a dataset of noise-free and regularly spaced observations with $\dt=0.1$. Top row illustrates the mean reward curves (computed for $H=30$ seconds) and 90/10 quantiles over rounds and the predictive mean squared error of the dynamics on a test set is visualized over rounds on the bottom row. We observe that all models perform similarly despite the discrepancy in the dynamics accuracy.}
\label{fig:comparison5}
\end{figure*}

\begin{figure}[b]
    \centering
    \includegraphics[width=.49\linewidth]{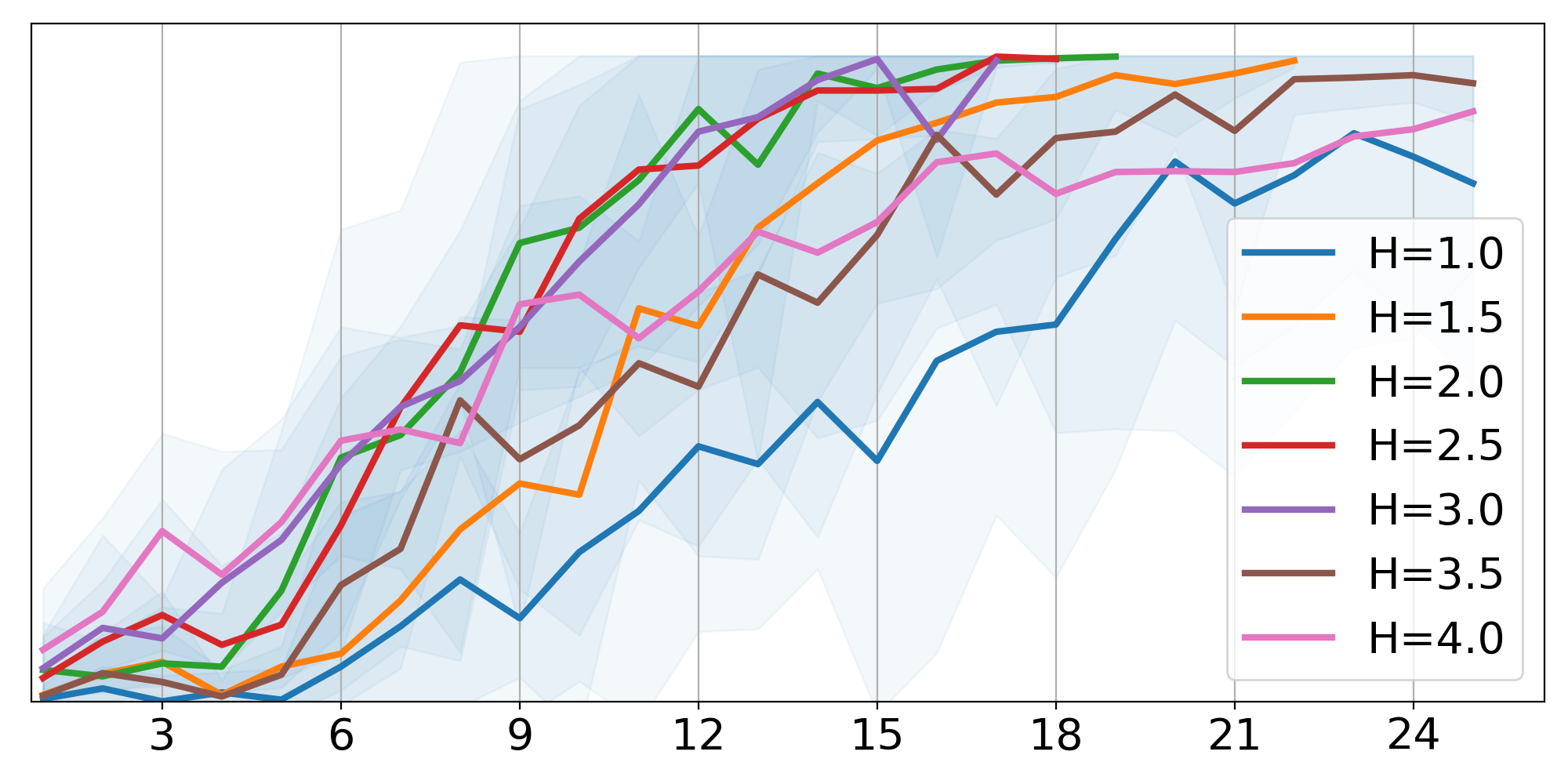}
    \includegraphics[width=.49\linewidth]{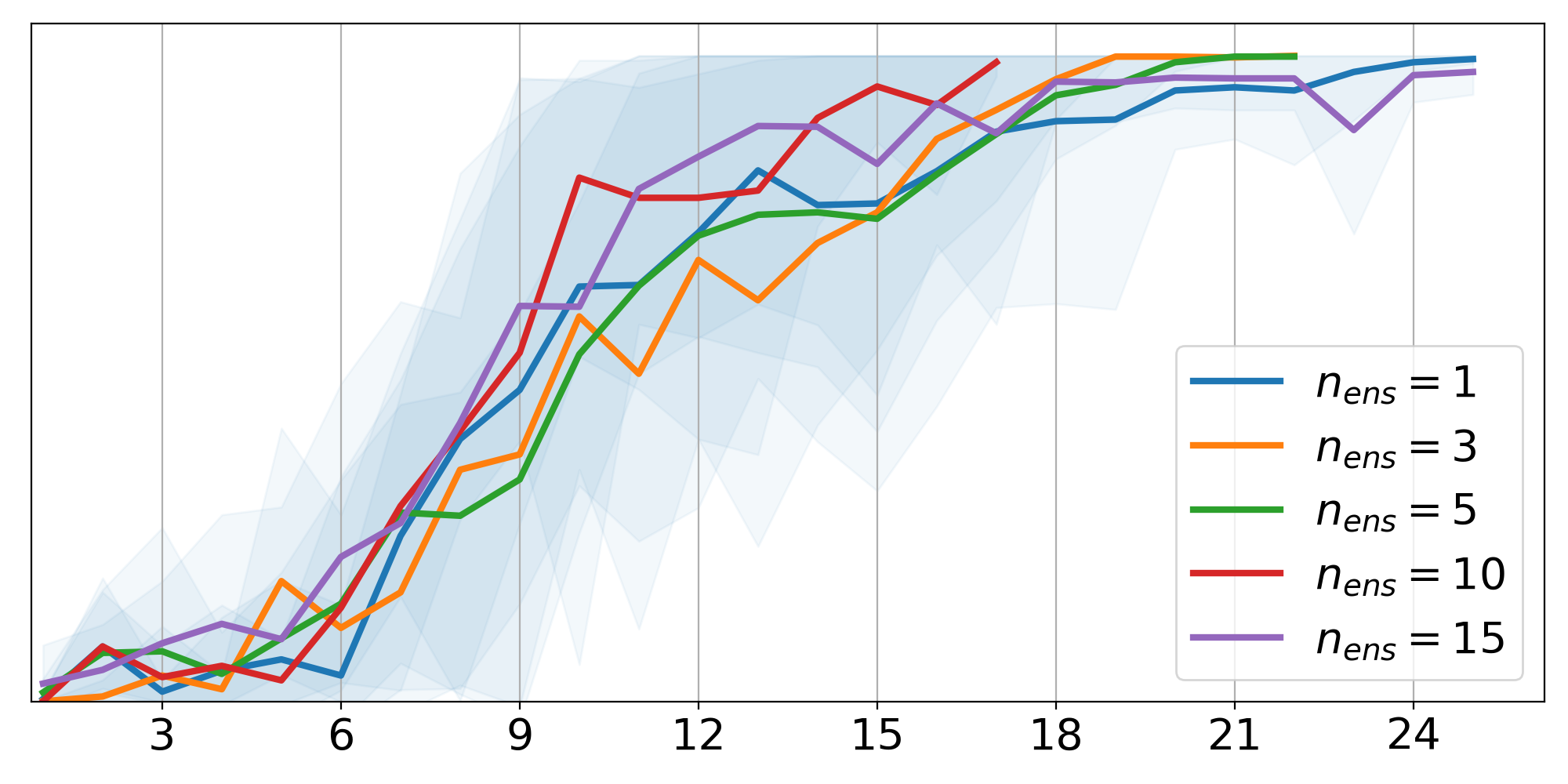}
    \caption{Ablation studies showing our ENODE approximation's sensitivity to the imagination horizon $H$ and the number of ensemble members $n_\text{ens}$.}
    \label{fig:ablations}
\end{figure}

\begin{figure*}[t]
    \includegraphics[width=.33\linewidth]{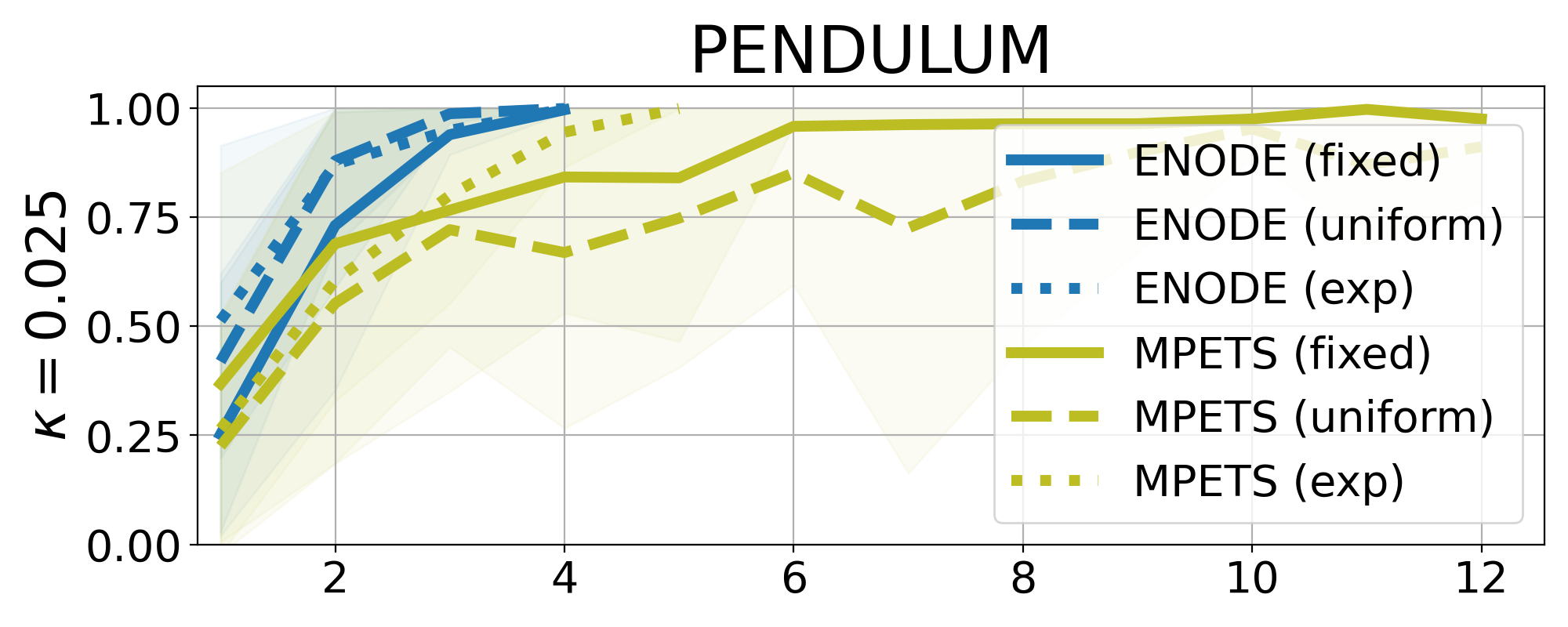}
    \includegraphics[width=.33\linewidth]{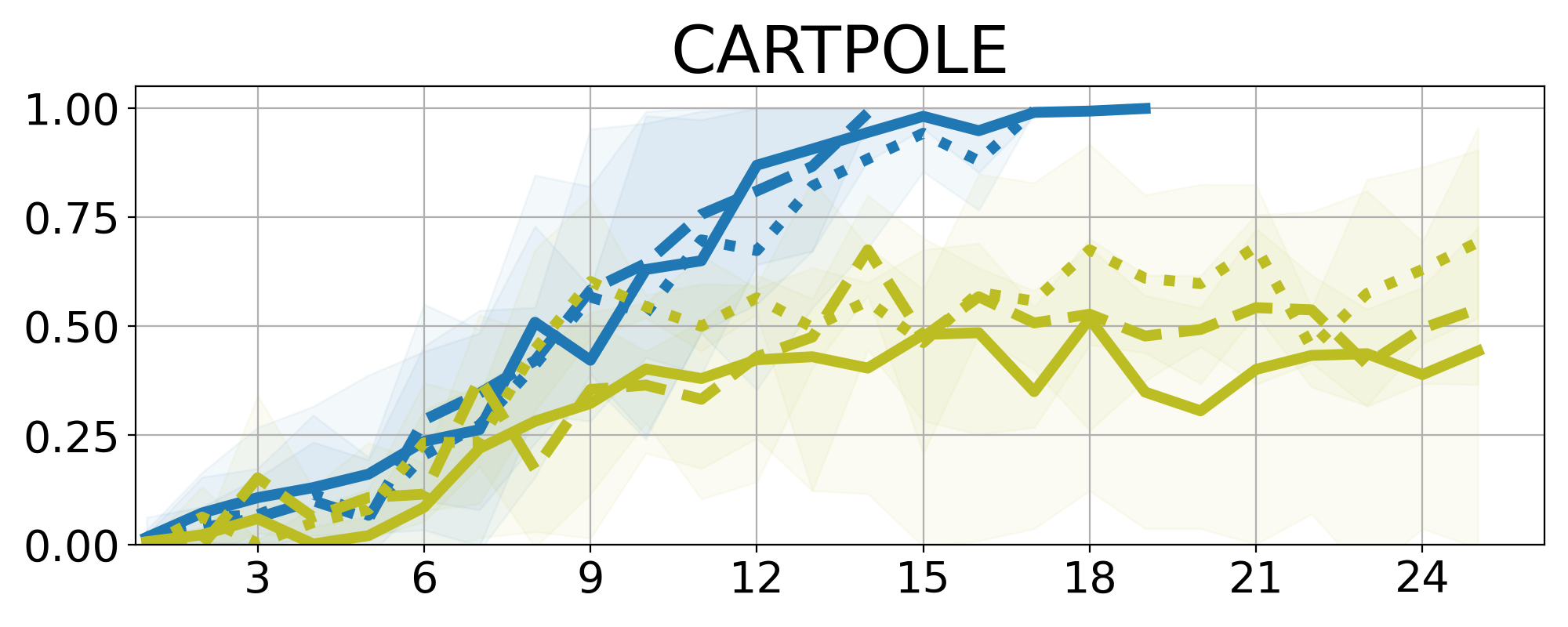}
    \includegraphics[width=.33\linewidth]{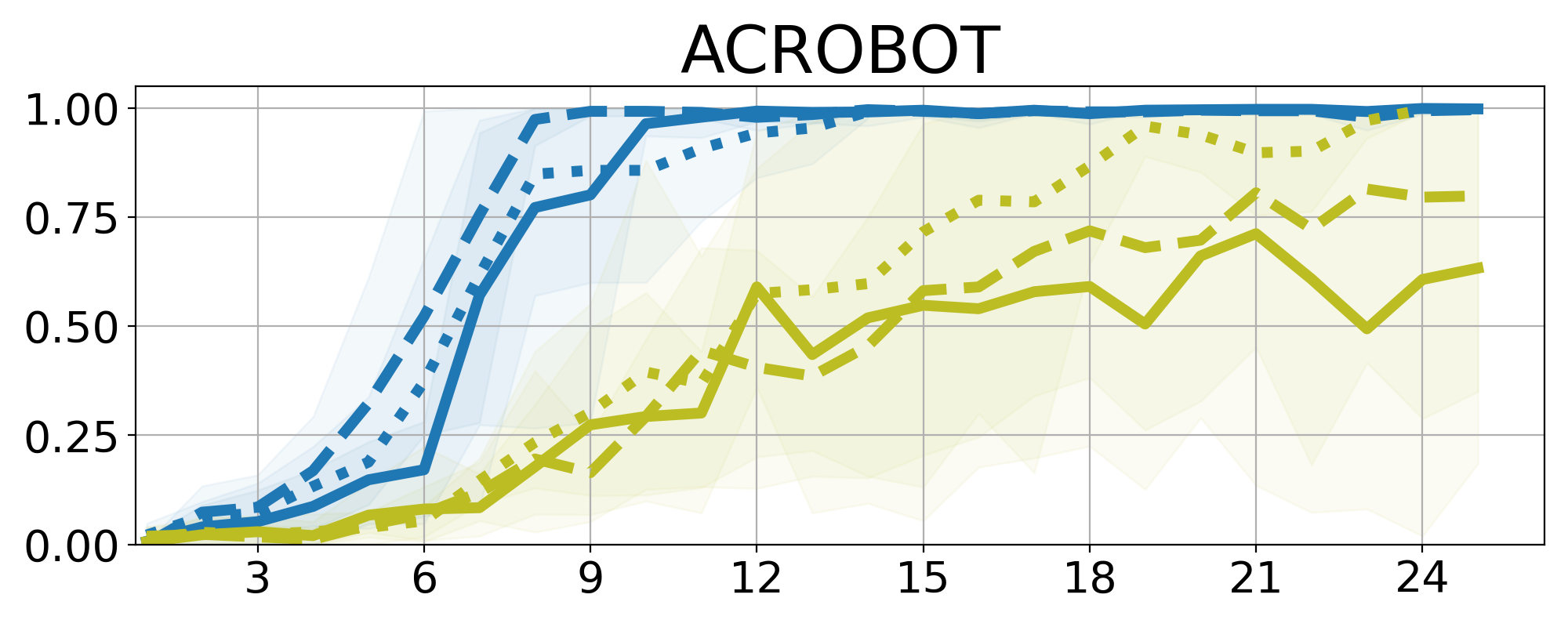} \\
    \includegraphics[width=.33\linewidth]{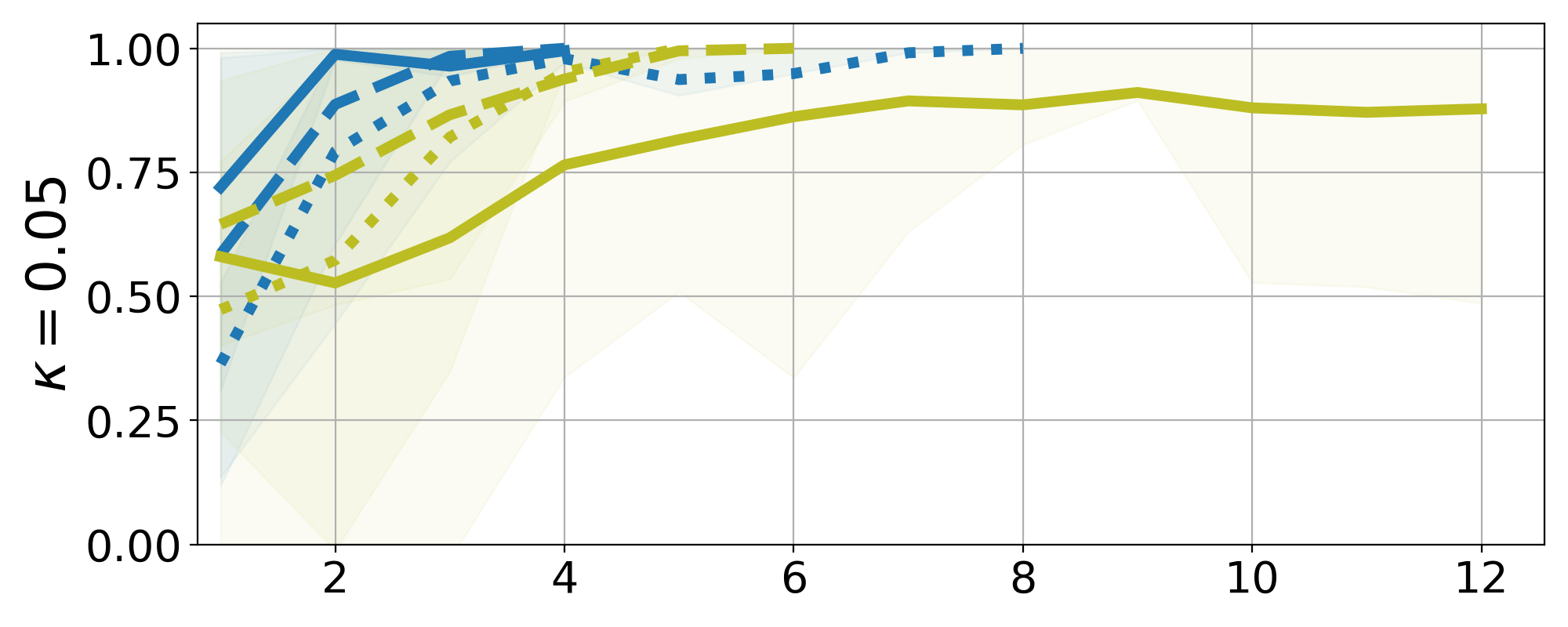}
    \includegraphics[width=.33\linewidth]{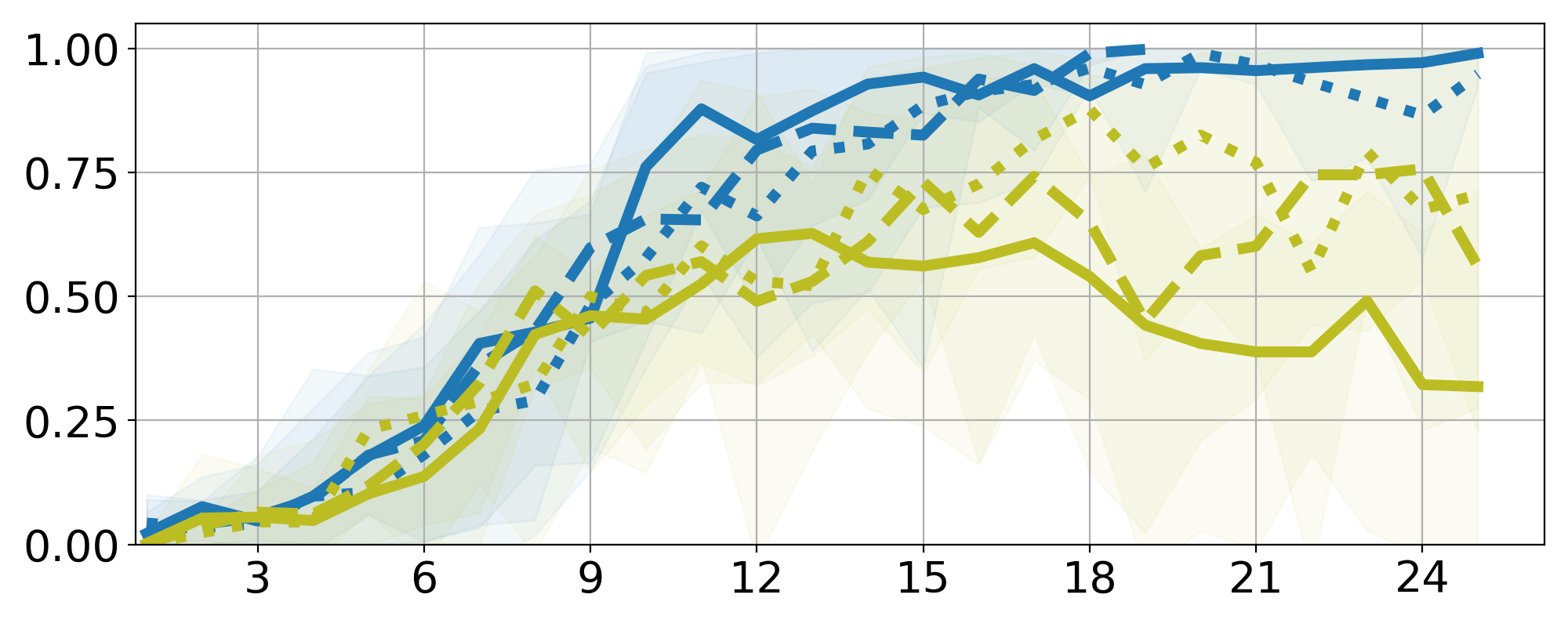}
    \includegraphics[width=.33\linewidth]{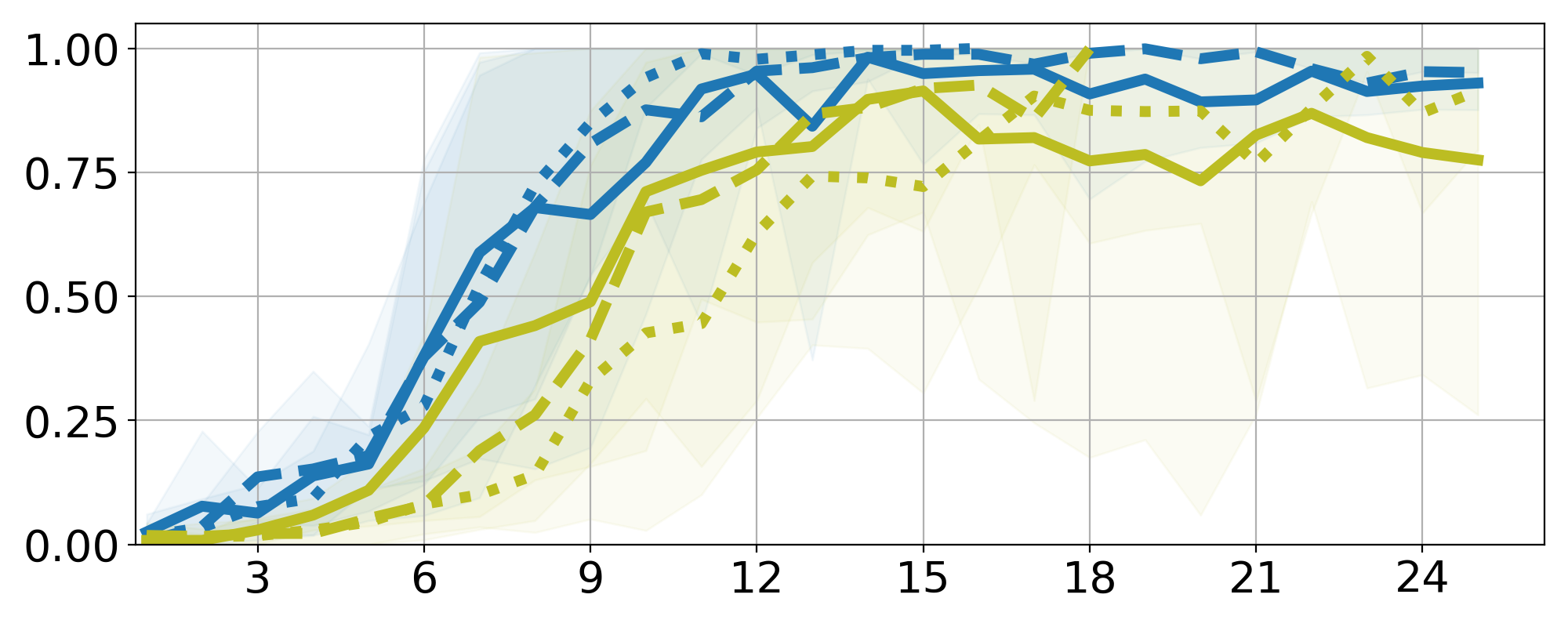} \\
    \includegraphics[width=.33\linewidth]{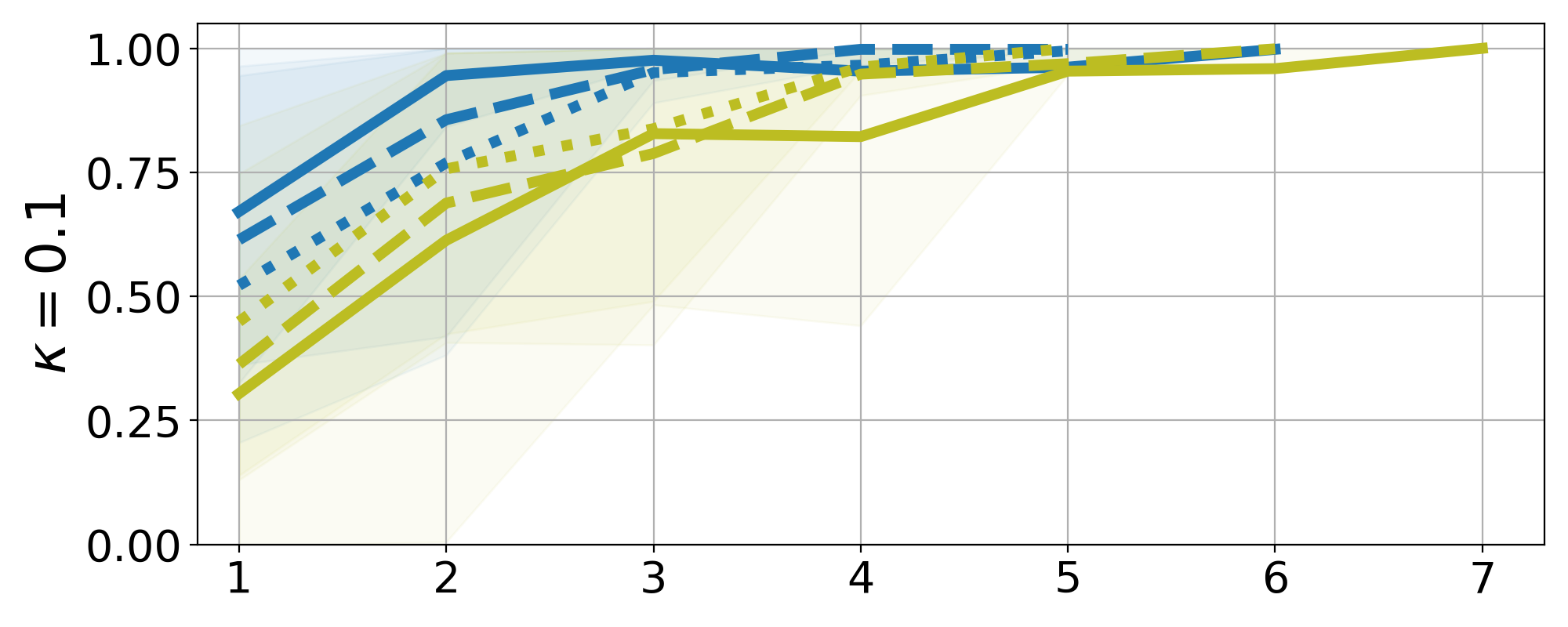}
    \includegraphics[width=.33\linewidth]{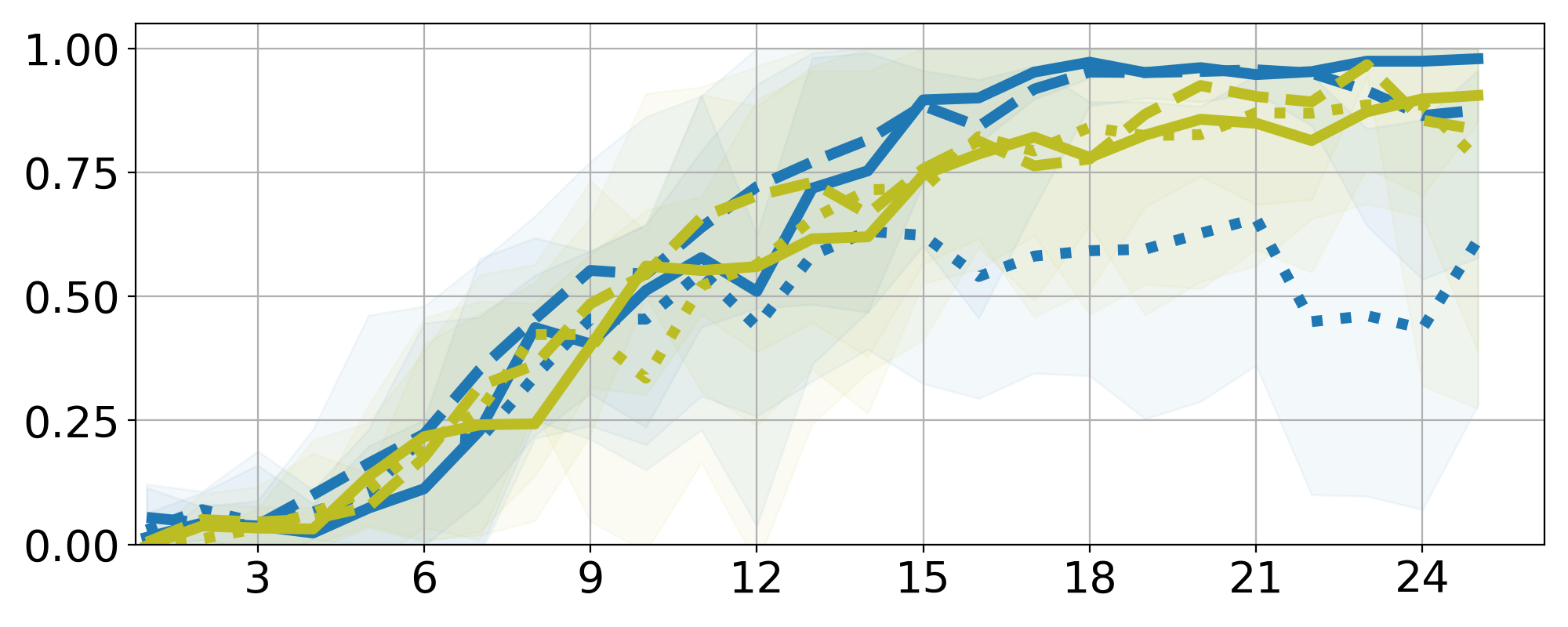}
    \includegraphics[width=.33\linewidth]{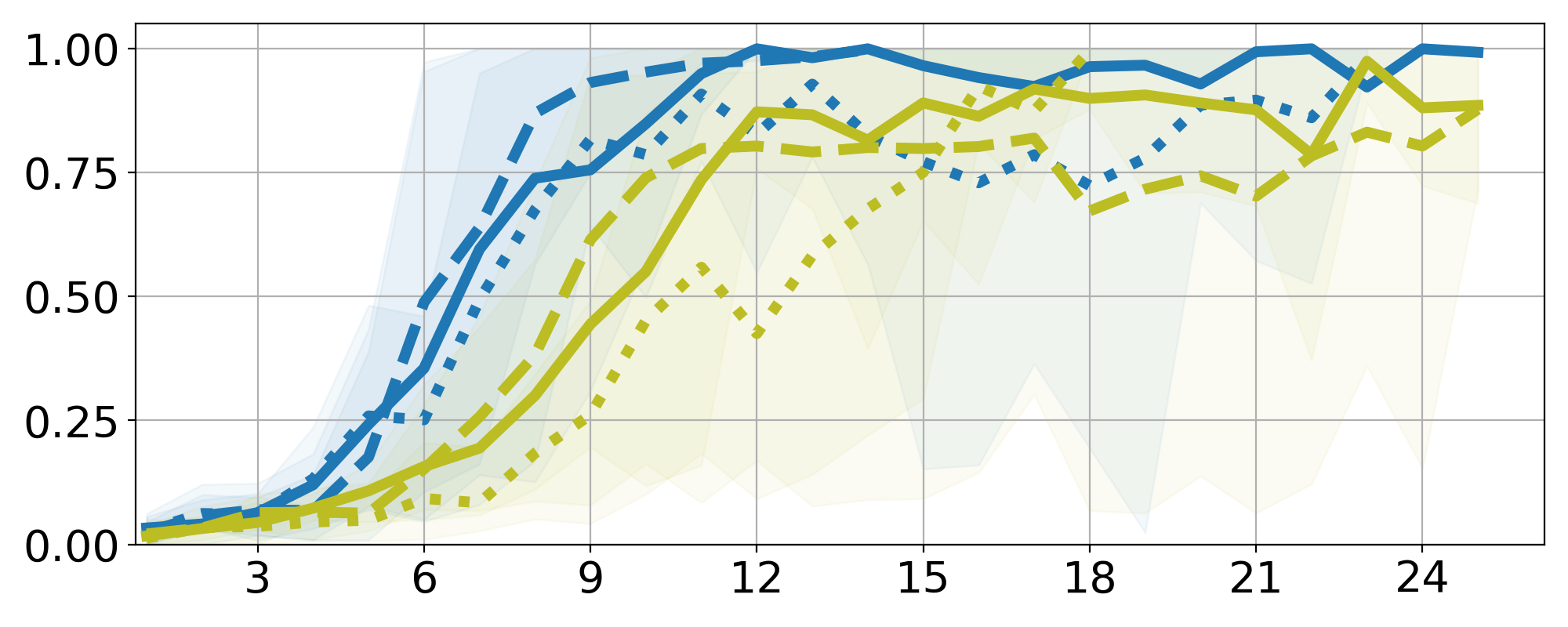}
    \\
    \includegraphics[width=.33\linewidth]{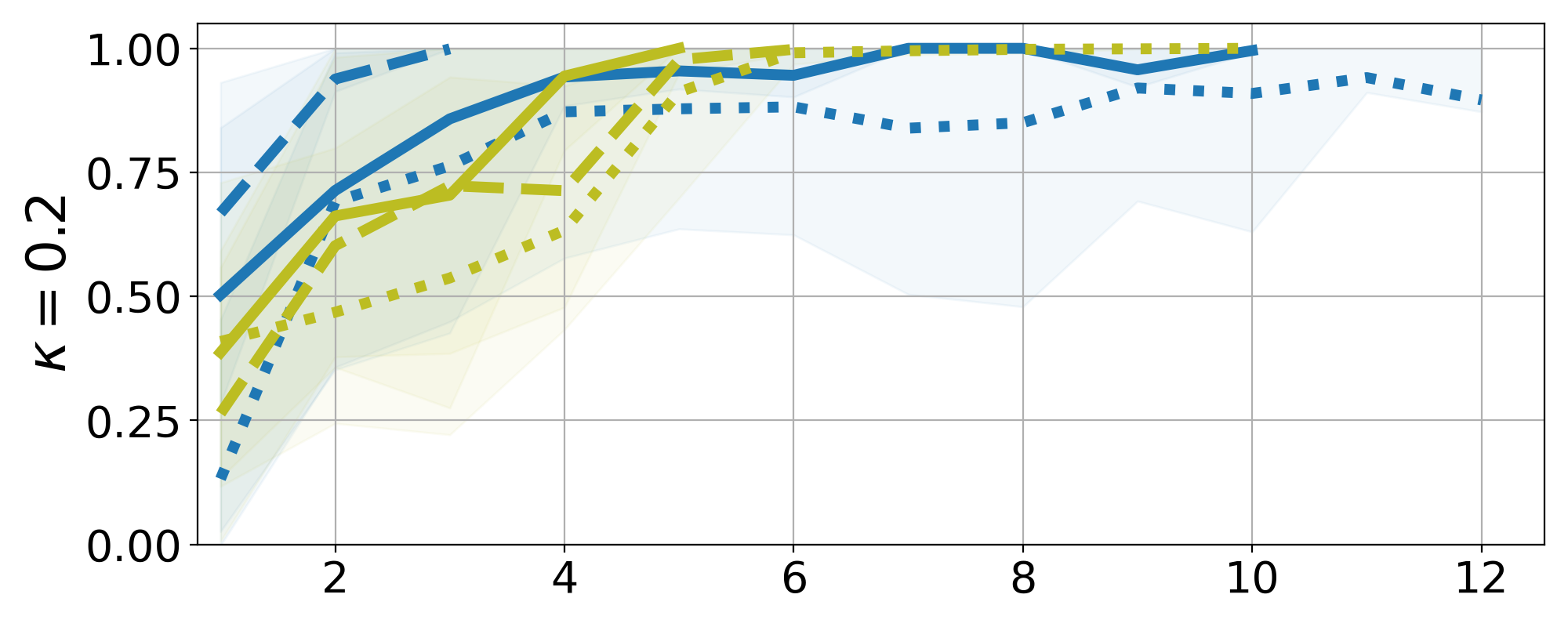}
    \includegraphics[width=.33\linewidth]{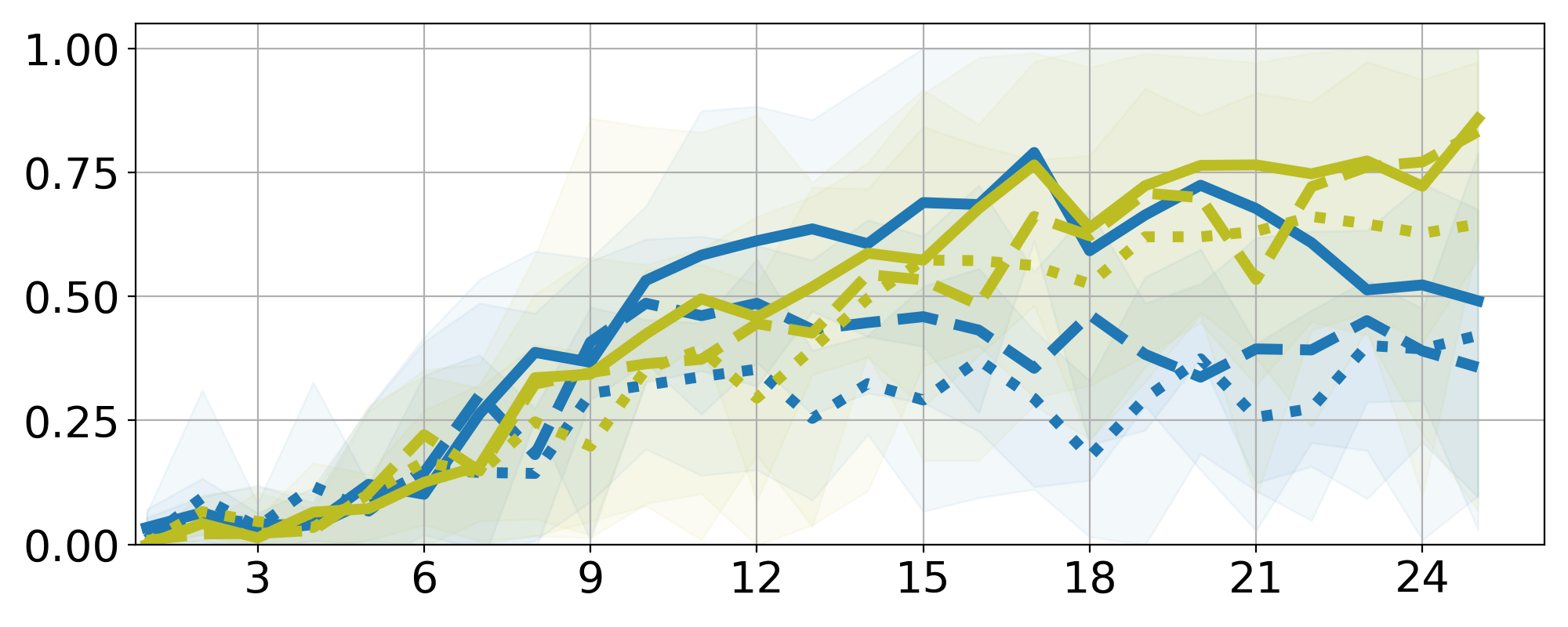}
    \includegraphics[width=.33\linewidth]{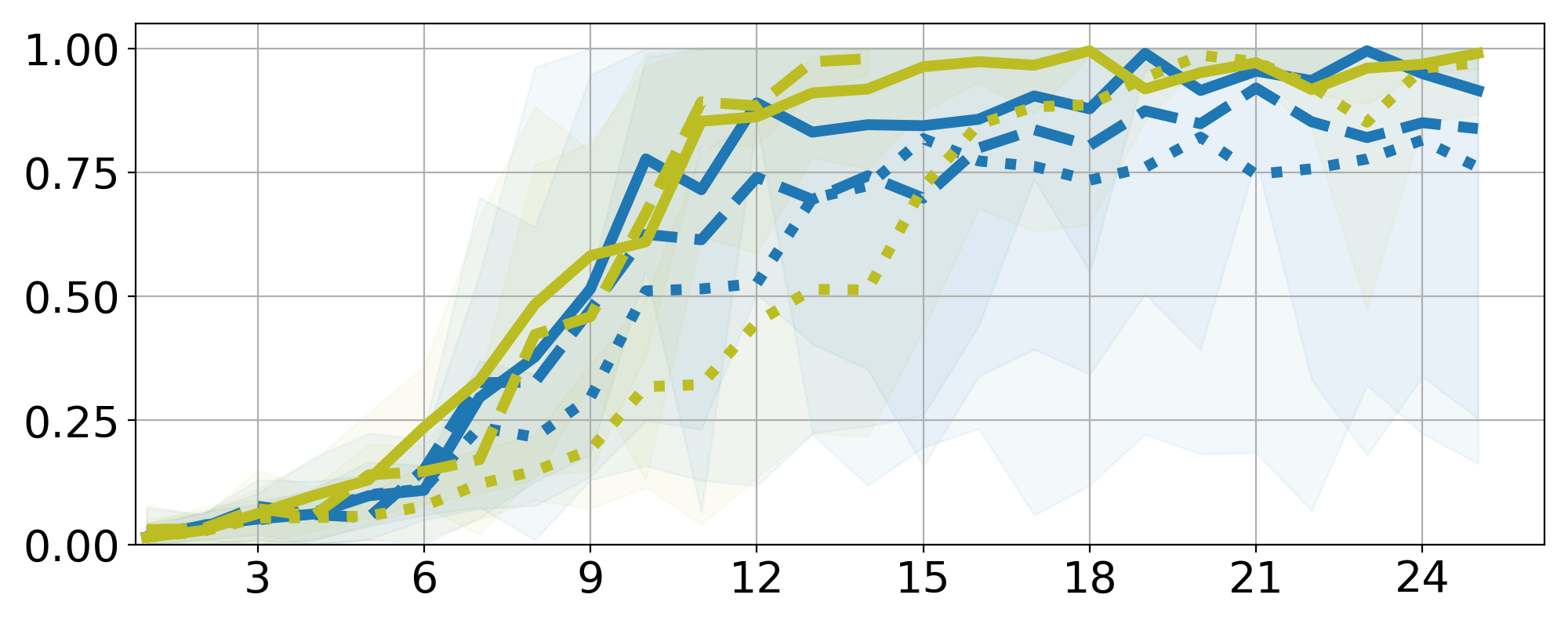}
\caption{A comparison of ENODE and MPETS on three noisy environments ($\sigma=0.025$) with three different observation spacings and varying mean time differences $\kappa$ (note that all three spacings have the same mean time increments $\kappa$). We observe that MPETS' performance deteriorates with reduced $\dt$ while our method is more robust to these changes.}
\label{fig:comparison2}
\end{figure*} 

\subsection{Vanilla vs.\  Modified PETS}
In this experiment, we investigate whether vanilla PETS can accurately infer the dynamics when the data sequences are temporally irregular. For this, both versions of PETS are trained on two CartPole datasets with observations arriving at fixed ($\Dt=0.1$) or uniformly sampled ($\Dt\sim \U(0,0.2]$) increments. The train and test one-step ahead prediction errors are illustrated in Figure~\ref{fig:pets-comp}. Both models successfully learn the transitions when they arrive at fixed increments. Modified PETS achieves near-zero error on temporally irregular dataset as well whereas the vanilla version cannot perfectly fit the training data as conjectured in Section~\ref{sec:dyn}. We proceed with modified PETS (MPETS) in our irregular sampling experiments.

\subsection{ENODE Hyperparameter Sensitivity}
As an ablation study, we check how the imagination horizon $H$ and the number of ensemble members $n_\text{ens}$ impact the model performance. We consider a noise-free CartPole environment with fixed observation time increments ($\Dt=0.1$), and disjointly vary $H \in [1.0,4.0]$ and $n_\text{ens} \in \{1,3,5,10,15\}$. The mean reward values over rounds are plotted in Figure~\ref{fig:ablations}. We observe that our ENODE approximation is somewhat robust to $n_\text{ens}$. As expected, too short or long horizons deteriorate the performance. In the remainder of this paper, we fix $H=2.0$ and $n_\text{ens}=10$.

\subsection{Comparisons on Standard Setup} 
We first compare our model variants, Deep PILCO and PETS on Pendulum, CartPole and Acrobot environments with noise-free and regularly sampled observations ($\Dt=0.1$). The top tow in Figure~\ref{fig:comparison5} illustrates the mean values obtained by executing policies in the real world. We observe that all our model variants as well as PETS perform comparably while Deep PILCO requires more rounds to solve the problems.

Next, we examine how the model performance is related to the dynamics accuracy. For this, we take a snapshot of all models after each round, and also collect a separate test dataset by executing the policies that solve the problems in the real world. The bottom panels in Figure~\ref{fig:comparison5} show how the dynamics errors evolve as Algorithm~\ref{alg:ctrl} proceeds. Specifically, we plot two-second ahead prediction error, which also equals to the imagination horizon in our Actor-Critic algorithm. Since PETS errors explode in early rounds, we clip it to 100 for visualization purposes. Our method attains high accuracy only after a few rounds whereas the policy converges to optimality much later. We conjecture that dynamics accuracy does not solely predict the overall performance. Additional results demonstrating shorter term predictions can be found in supplementary Figure S2.

\subsection{Noisy and Temporally Irregular Data Experiments}
To contrast the robustness of discrete and continuous-time techniques, we evaluate MPETS and ENODE on more realistic datasets of irregularly sampled and noisy data sequences, where the observations are perturbed by Gaussian noise with standard deviation $\sigma=0.025$. We also vary the mean observation time differences $\kappa \in [0.025, 0.05, 0.1, 0.2]$ for all three observation spacings.

Figure~\ref{fig:comparison2} exhibits mean value curves on three tasks of our interest. The most striking observation is that small time differences lead to the breakdown of MPETS whereas our model is more tolerant. We believe MPETS' performance drop is inevitable since noise corrupts both the inputs and outputs of the transition function. On the other hand, our method infers the underlying process and thus noise can be handled by the observation noise model without corrupting the dynamics inference. Furthermore, the performances of both models stay roughly unaffected with stochastic time increments. We conclude by highlighting the fact that both models solve the tasks in fewer rounds when the dataset is noise-free and the models perform comparably well.

\section{Discussion}
We have presented a novel continuous-time MBRL framework that resolves the time discretization approximation of existing techniques. Our dynamics learning method relies on the neural ODE framework, and thus can learn arbitrarily complicated controlled ODE systems. We propose several original strategies to handle epistemic uncertainty in NODE. Furthermore, we develop a new actor-critic algorithm that operates in continuous time and space. We experimentally demonstrate that our method is robust to environment changes such as observation noise, temporal sparsity and irregularity.


\vfill
\pagebreak

\newpage

\setcounter{section}{0}
\setcounter{equation}{0}
\setcounter{figure}{0}
\setcounter{table}{0}
\setcounter{page}{1}
 \renewcommand{\thesection}{S\arabic{section}}
 \renewcommand{\theequation}{S\arabic{equation}}
 \renewcommand{\thefigure}{S\arabic{figure}}

\twocolumn[
\icmltitle{Continuous-Time Model-Based Reinforcement Learning \\ {\normalsize SUPPLEMENTARY MATERIAL}}





\vskip 0.3in
]

\section{Technical Discussion}

\subsection{The Form of the Differential Function}
For notational convenience, we defined the problem as a first-order ODE system:
\begin{equation*}
    \dot{\s}(t) = \frac{d\s(t)}{dt} = \f(\s(t),\a(t)).
\end{equation*}
We now explain possible extensions to this formulation to inject problem-specific prior knowledge. 

\paragraph{Linearity w.r.t. action}
In many Newtonian systems, the action affects the time differential linearly, a widespread assumption in CT control literature \cite{doya2000reinforcement}. Estimating such systems with an arbitrary function of state and action would tie the action and the dynamics in a nonlinear way, which would render the learning problem unnecessarily complicated. Therefore, we propose to decompose the differential function as follows \cite{vamvoudakis2010online}:
\begin{equation*}
\frac{d \s(t)}{dt} = \f(\s(t))+ \h(\s(t)) \cdot \a(t). \label{eq:lin-act}
\end{equation*}
where $\h : \R^{d} \mapsto \R^{d \times m}$ and $\cdot$ denotes the standard matrix-vector product. Above formulation assumes an additive differential function of \textit{dynamics} and \textit{control} component. The former aims to learn the evolution of the system under zero force whereas $\h$ defines a manifold the action is projected onto. Since we approximate these functions with neural networks, both the dynamics and the manifold can be arbitrarily complicated.

\paragraph{Second-order dynamics}
Most dynamical systems can be expressed in terms of position $\s \in \R^d$ and velocity $\v \in \R^d$ components. Such decomposition of the state space is shown to better explain the phenomena of interest if the underlying physics is indeed second-order \cite{yildiz2019ode2vae}. Formally, a second-order dynamical system with control is defined as follows:
\begin{equation*}
    \frac{d \s(t)}{d t} = \v(t), \qquad \frac{d \v(t)}{d t} = \f(\s(t),\v(t),\a(t)).
\end{equation*}
Here, $\f: \R^{2d+m}\rightarrow\R^d$ is referred to as \textit{acceleration field}.

\paragraph{Hamiltonian dynamics}
\citet{zhong2019symplectic} already describes Hamiltonian dynamics in RL context very clearly; however, we include this subsection for completeness. Hamiltonian mechanics reformulate classical physical systems in terms of canonical coordinates $(\q, \p)$ with $\q,\p \in \R^d$, where $\q$ denotes generalized coordinates and $\p$ is their conjugate momenta. The time evolution of a Hamiltonian system is defined as
\begin{equation*}
    \frac{d\q}{dt} = \frac{d\H}{d\p}, \qquad \frac{d\p}{dt} = - \frac{d\H}{d\q} \label{eq:ham}
\end{equation*}
where $\H(\q,\p): \R^{2d} \rightarrow \R$ denotes the Hamiltonian. For simplicity, we assume a time-invariant Hamiltonian, which also corresponds to the total energy of the system. Typically, Hamiltonian is decomposed into a sum of kinetic energy $T$ and potential energy $V$:
\begin{equation*}
    \H=T+V, \qquad T=\frac{\p^T M^{-1}(\q) \p}{2m}, \qquad V=V(\q)  \label{eq:ham2}
\end{equation*}
where $m$ denotes the mass. In simple systems such as pendulum, the mass matrix $M(\q)$ is an identity matrix, implying Euclidean geometry. More complicated systems like cart-pole requires learning the geometry. Given a Hamiltonian decomposing like above, the dynamics become
\begin{equation*}
    \frac{d\q}{dt} = \frac{M^{-1}(\q) \p}{m}, \qquad \frac{d\p}{dt} = - \frac{dV}{d\q}  \label{eq:ham3}
\end{equation*}
The dynamics learning problem reduces to learning a potential energy function $\V(\q): \R^d \rightarrow \R$, whose derivative gives the time evolution of momentum, and estimating the geometry through its Cholesky decomposition $L$, i.e., $LL^{-1}=M$, via an additional neural network.

\subsection{Greedy policy}
A greedy policy is defined as the one that minimizes the Hamilton–Jacobi–Bellman equation:
\begin{equation}
    V^*(\s) = \min_\a \left[ \frac{dV(\s)}{d\s} \cdot \f(\s_t,\a) + r(\s,\a)  \right]  \label{eq:hjb}
\end{equation}
The greedy policy can be expressed in closed form if \textit{(i)} the reward is of the form $r(\s,\a)=r_\s(\s)+r_\a(\a)$ with an invertible action reward and \textit{(ii)} the system dynamics is linear with respect to the action as in eq. \eqref{eq:lin-act} \cite{doya2000reinforcement,tassa2007least}:
\begin{equation}
    \a^*_t = -\frac{dr_\a^{-1}}{d\a} \left( \frac{\f(\s_t,\a)}{d\a}^T \frac{dV(\s)}{d\s}^T \right) \label{eq:opt-act}
\end{equation}
Consequently, given the above assumptions are satisfied, the optimal policy can be expressed in closed form for a given value function, which would obviate the need for an actor. We leave the investigation of model-based greedy policy as an interesting future work. 

\section{Experiment Details}
This section consists of experimental details which are not included in the main text. 

\subsection{Environments}
The environment-specific parameters are given in Table \ref{tab:envs}. In all environments, the actions are continuous and restricted to a range $[-a_\text{max},a_\text{max}]$. Similar to \citet{zhong2019symplectic}, we experiment with the fully actuated version of the Acrobot environment since no method was able to solve the under-actuated balancing problem. 

\subsection{Reward Functions}
Assuming that each observation $\s=(\q,\p)$ consists of state (position) $\q$ and velocity (momentum) $\p$ components, the differentiable reward functions have the following form:
$$ r(\q,\p,\a) = \exp\left( -||\q-\s^\text{goal}||_2^2 - c_\p ||\p||_2^2 \right) - c_\a ||\a||_2^2 $$
where $c_\p$ and $c_\a$ denote environment-specific constants. The exponential function aims to restrict the reward in a range [0,1] minus the action cost, which aids learning \citep{deisenroth2011pilco}. The constants $c_\p$ and $c_\a$ are set so that they \textit{(i)} penalize large values, and \textit{(ii)} do not enforce the model to stuck at trivial local optima such as the initial state. 

\paragraph{Goal states} The goal state $[0,\ell]$ in Pendulum environment corresponds to $x$ and $y$ coordinates of pole's tip, with $\ell$ being the length of the pole. The reward is maximized when the pole is fully upright. In CartPole, this state is concatenated with 0, which represents the cart's target location. Finally in Acrobot, the goal state involves the $x$ and $y$ coordinates of the second link only (hence $2D$).  

\subsection{Dataset}
\paragraph{Initial dataset} Each experiment starts with collecting an initial dataset of $N_0$ trajectories at observation time points with length $T=50$. In all environments, the initial state is distributed uniformly: 
$$\s_0 \sim \mathcal{U}[-\s^\text{box},\s^\text{box}].$$  
Initial random actions are drawn from a Gaussian process:
$$\a_{t} \sim \mathcal{GP}(\boldsymbol{0},K(t,t')).$$
The inputs to the GP are the observation time points. We opt for a squared exponential kernel function with $\sigma=0.5$ and $\ell=0.5$. The output of the GP is followed by a TANH function and multiplied with $a_{\text{max}}$.

\paragraph{Data collection} After each round, the policy is executed once in the environment starting from an initial state drawn from the environment. This is followed by the execution of $N_{\text{exp}}$ exploring policy functions. Similar to the idea of perturbing policies with an Ornstein–Uhlenbeck process for exploration \citep{lillicrap2015continuous}, we add draws from a zero-mean Gaussian process to the policy for smoother perturbations:
\begin{align*}
    \bpi^{\text{explore}}\left(\s(t),t\right) &:= \bpi\left(\s(t)\right)+\z(t) \\
    \z(t) &\sim \mathcal{GP}(\boldsymbol{0},k(t,t'))
\end{align*}
where the input to the GP is a set of time points. We again use a squared exponential kernel function with $\sigma=0.1$ and $\ell=0.5$. We choose the initial values for the exploring data sequences from the previous experience dataset randomly, where each state has a weight propotional to the dynamics estimator's variance at that state.

\subsection{Training Details}
We used ADAM optimizer to train all the model components \citep{kingma2014adam}. More explanation is as follows: 
\begin{itemize}
\itemsep0em 
    \item \textbf{NODE Dynamics:} We initialize the differential function by gradient matching:
    \begin{equation*}
        \f(\s_i,\a_i) \approx \frac{\s_{i+1}-\s_{i}}{\t_{i+1}-t_i}
    \end{equation*}
    Regardless of how observation time points are distributed, we use subsequences of length $t_s=5$ to train the dynamics model. We randomly pick 5 subsequences from each data trajectory to reduce gradient stochasticity. While training the neural ODE model, we start with an initial learning rate of 1e-4, gradually increase it to 1e-3 in 100 iterations, and then proceed $N_\text{dyn}=1250$ iterations with the latter learning rate.
    \item \textbf{Discrete Dynamics:} We train PETS and deep PILCO with the algorithms given respective papers. 
    \item \textbf{Actor-Critic:} In each round, we form a dataset of initial values from the experience dataset. We chose to exclude the sequences collected with exploring policy as they may explore states that are undesirably far from the goal. We set $N_\text{ac}=250$.  
\end{itemize}

\begin{table*}[ht]
	\caption{Environment specifications}
	\label{tab:envs}
	\vskip 0.15in
	\begin{center}
    \begin{small}
	\begin{tabular}{ccccccccc}
    	\toprule
		 Environment & $N_0$ & $N_{\text{exp}}$ & $c_\p$ & $c_\a$  & $a_{\text{max}}$ & $\s^\text{box}$ & $\s^\text{goal}$ & Max. execution time (h)  \\ 
	    \midrule
		 \textsc{Pendulum} & 3 & - & 1e-2 & 1e-2 & 2 & $[\pi,3]$ & $[0,\ell]$ & 12 \\
		 \textsc{CartPole} & 5 & 2 & 1e-2 & 1e-2 & 3 & $[0.05, 0.05, 0.05, 0.05]$ & $[0,0,\ell]$ & 24 \\
		 \textsc{Acrobot}  & 7 & 3 & 1e-4 & 1e-2 & 4 & $[0.1, 0.1, 0.1, 0.1]$  & $[0,2\ell]$ & 24 \\
        \bottomrule
    \end{tabular}
    \end{small}
    \end{center}
    \vskip -0.1in
\end{table*}

\subsection{Neural Network Architectures}
The dynamics, actor and critic functions are approximated by multi-layer perceptrons. In all methods and environments, we used the same neural network architectures, which are given as follows:
\begin{itemize}
\itemsep0em 
    \item \textbf{Dynamics:} 3-hidden layers, 200 hidden neurons and ELU activations. We experimentally observed that dynamics functions with ELU activations tend to extrapolate better on test seqeunces. Therefore, training the dynamics model with an augmented dataset (after each round) becomes much more robust.
    \item \textbf{Actor:} 2-hidden layers, 200 hidden neurons and RELU activations. Based on the idea that optimal policies can be expressed as a collection of piece-wise linear functions, we opt for RELU activations. Neural network output goes into TANH activation and multiplied with $a_{\text{max}}$.
    \item \textbf{Critic:} 2-hidden layers, 200 hidden neurons and TANH activations. Since the state-value functions must be smooth, TANH activation is more suitable compared to other activations. We empirically observed critic networks with RELU activations easily explode outside the training data, which deteriorates the learning.
\end{itemize}

\subsection{Additional Results}
Predictive dynamic errors on shorter sequences are illustrated in Figure~\ref{fig:dyn}. We see that future MSEs are much lower compared to $H=2$ while they still cannot directly predict overall model performance $V(\s_0)$

\section{ODE Solver Comparison}
In this ablation study, we ask two questions in relation with the simulation environment and numerical integration: \textit{(i)} Which numerical ODE solver one should use, \textit{(ii)} To what extent our continuous framework differ from its discrete counterparts? To answer, we have built a simple experiment on CartPole environment where several ODE solvers are compared: three adaptive step solvers (\texttt{dopri5} (\texttt{RK45}), \texttt{RK23} and \texttt{RK12}), five fixed step solvers (\texttt{RK4} with 1/10 intermediate steps, and \texttt{Euler} with 10/100/1000 intermediate steps), as well as discrete transitions. Due to the lack of a closed-form ODE solution, \textit{true} ODE solutions are obtained by Runge-Kutta 7(8) solver, the numerical integrator which achieves the smallest local error to the best of our knowledge \citep{prince1981high}. 

\begin{figure}[h]
    \includegraphics[width=.95\linewidth]{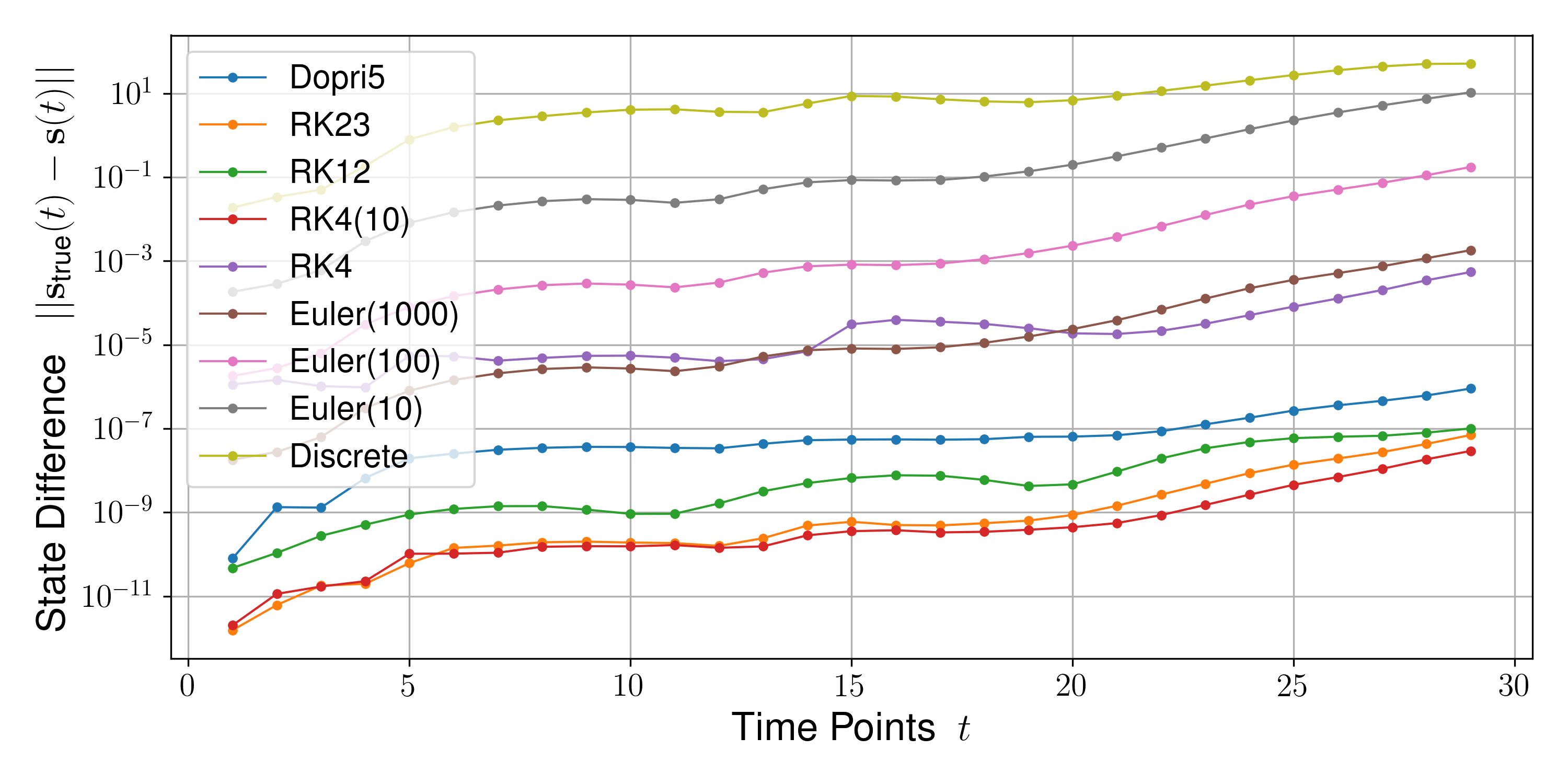}
    \caption{Error estimates of different numerical integration methods plotted against integration time.}
    \label{fig:ode-solver-comp}
\end{figure}

\begin{figure*}[h]
\hfill
\begin{minipage}[t]{0.33\textwidth}
    \centering
    \textsc{Pendulum}
    \includegraphics[width=\linewidth]{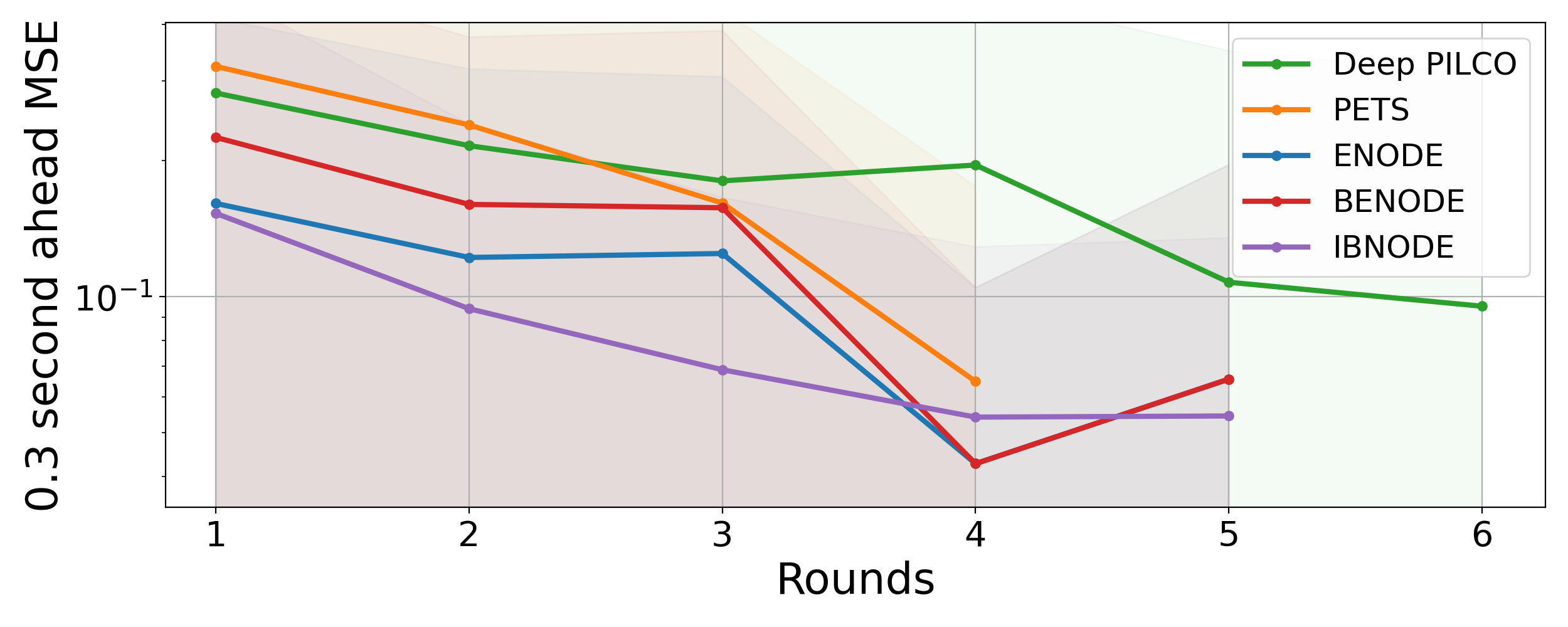}
\end{minipage}
\begin{minipage}[t]{0.33\textwidth}
    \centering
    \textsc{CartPole}
    \includegraphics[width=\linewidth]{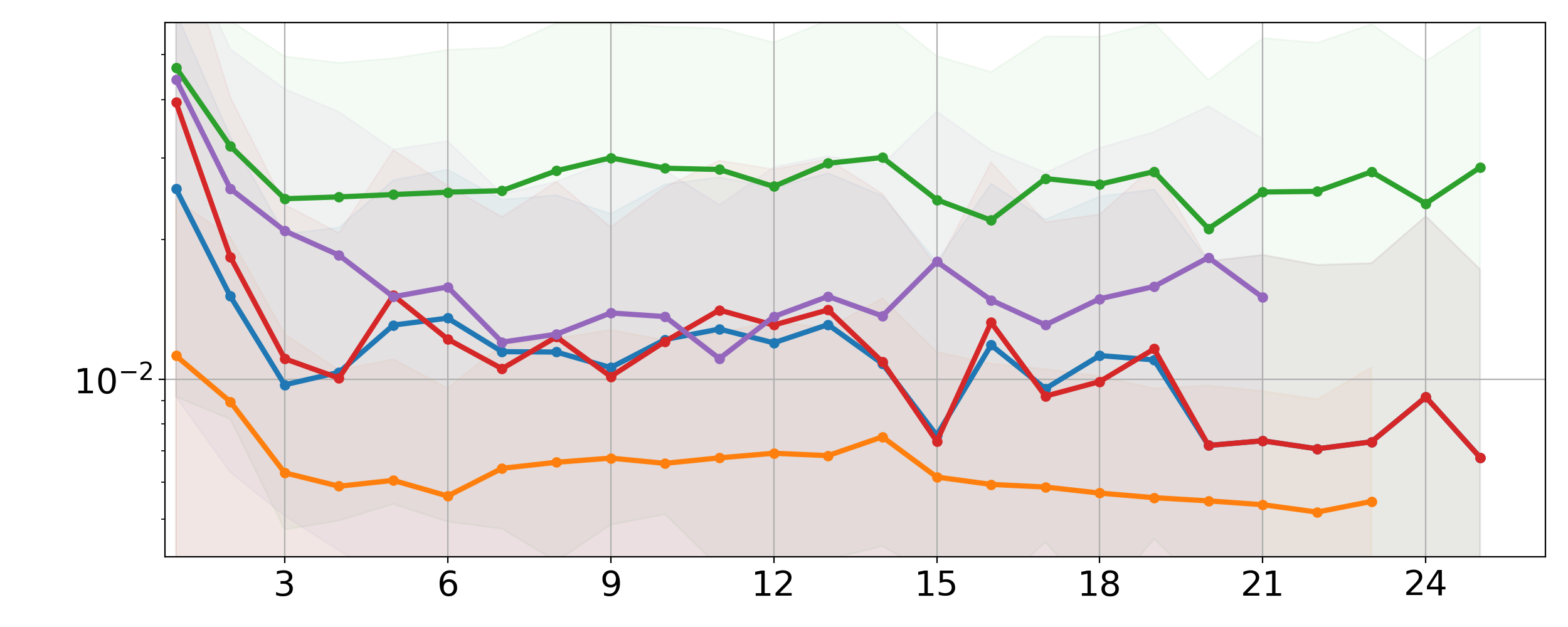}
\end{minipage}
\begin{minipage}[t]{0.33\textwidth}
    \centering
    \textsc{Acrobot}
    \includegraphics[width=\linewidth]{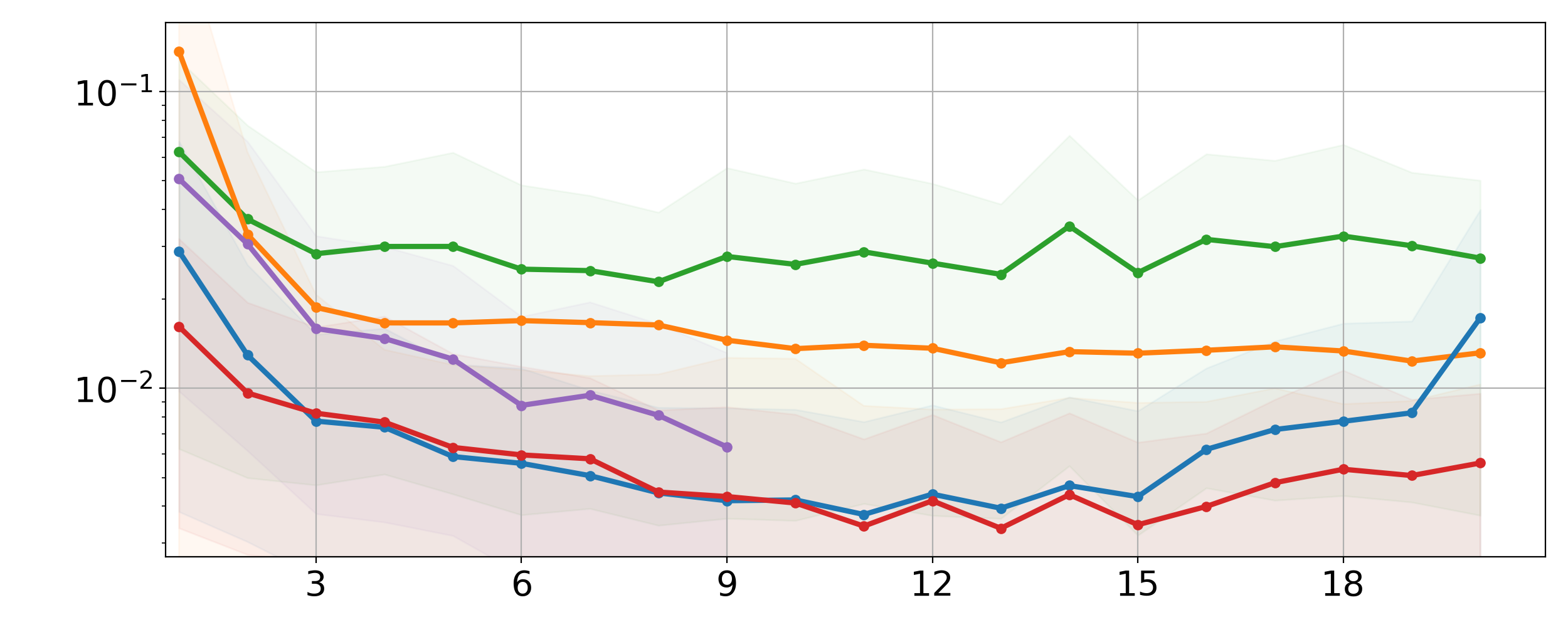}
\end{minipage}
\hfill
\begin{minipage}[t]{0.33\textwidth}
    \centering
    \includegraphics[width=\linewidth]{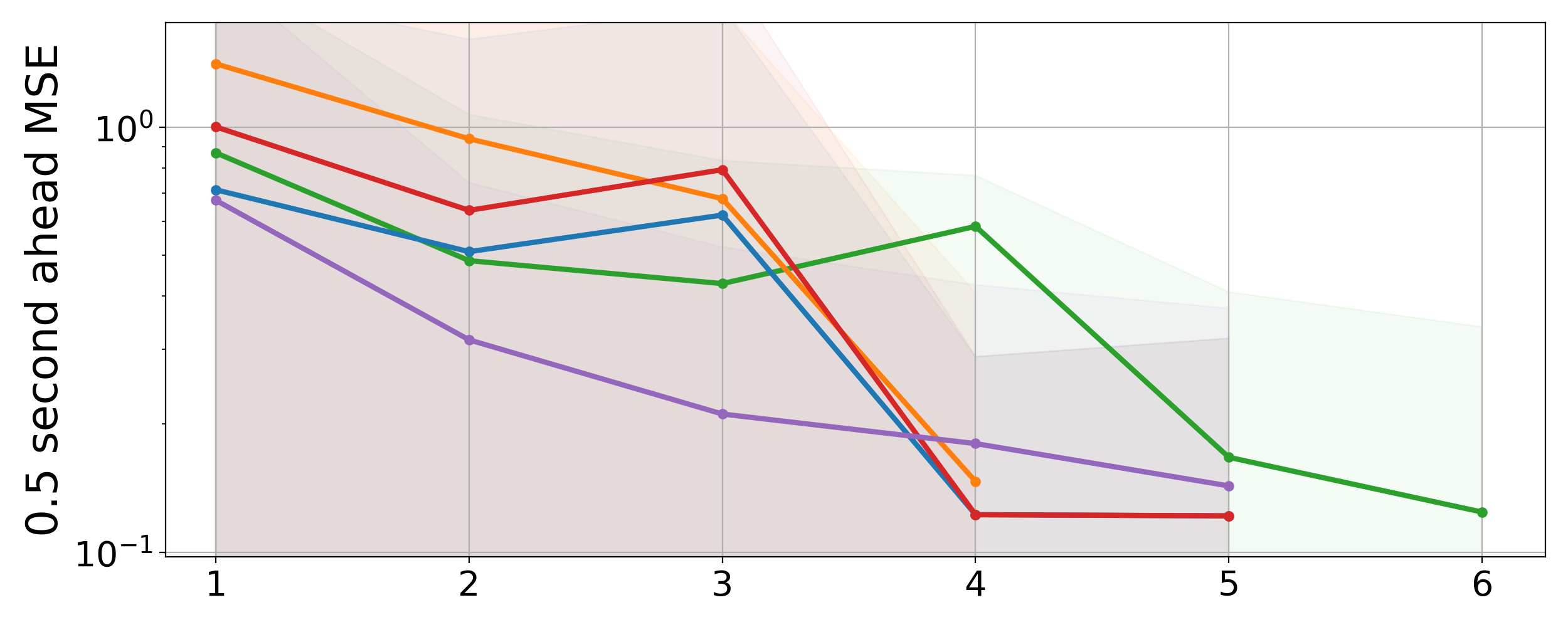}
\end{minipage}
\begin{minipage}[t]{0.33\textwidth}
    \centering
    \includegraphics[width=\linewidth]{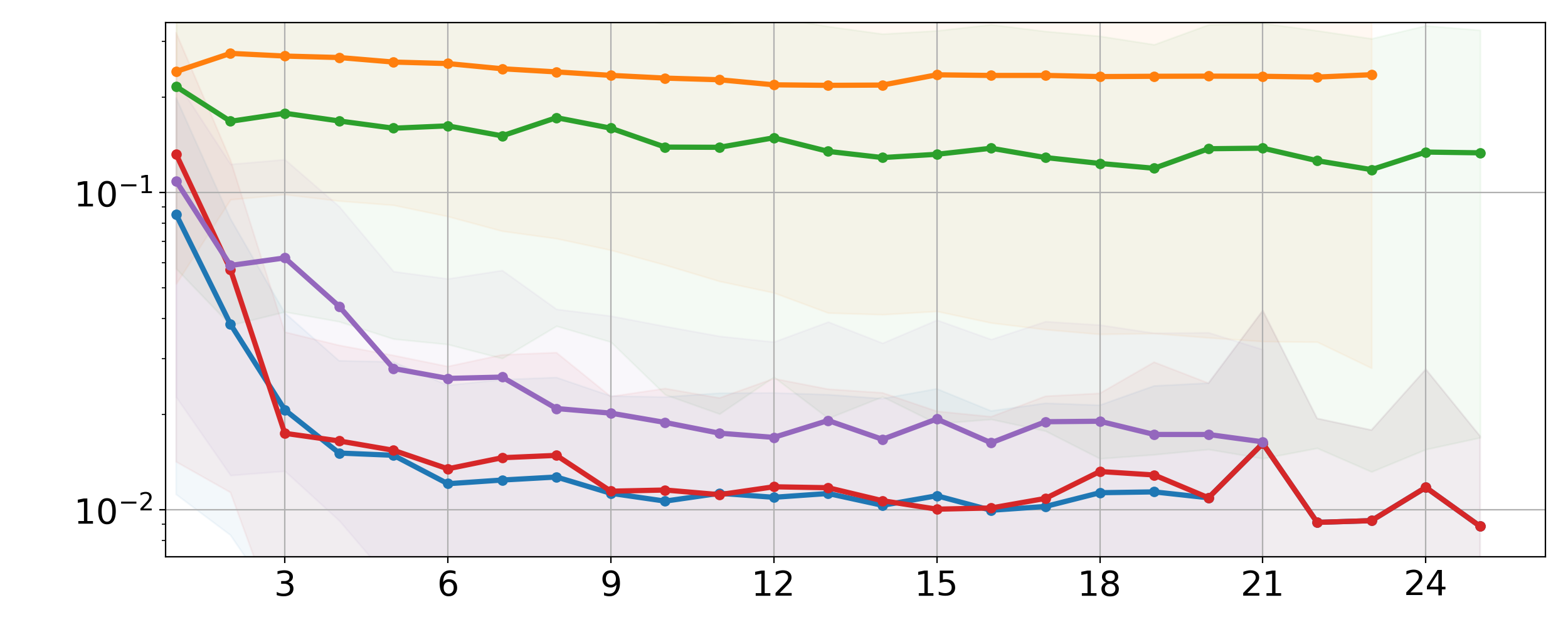}
\end{minipage}
\begin{minipage}[t]{0.33\textwidth}
    \centering
    \includegraphics[width=\linewidth]{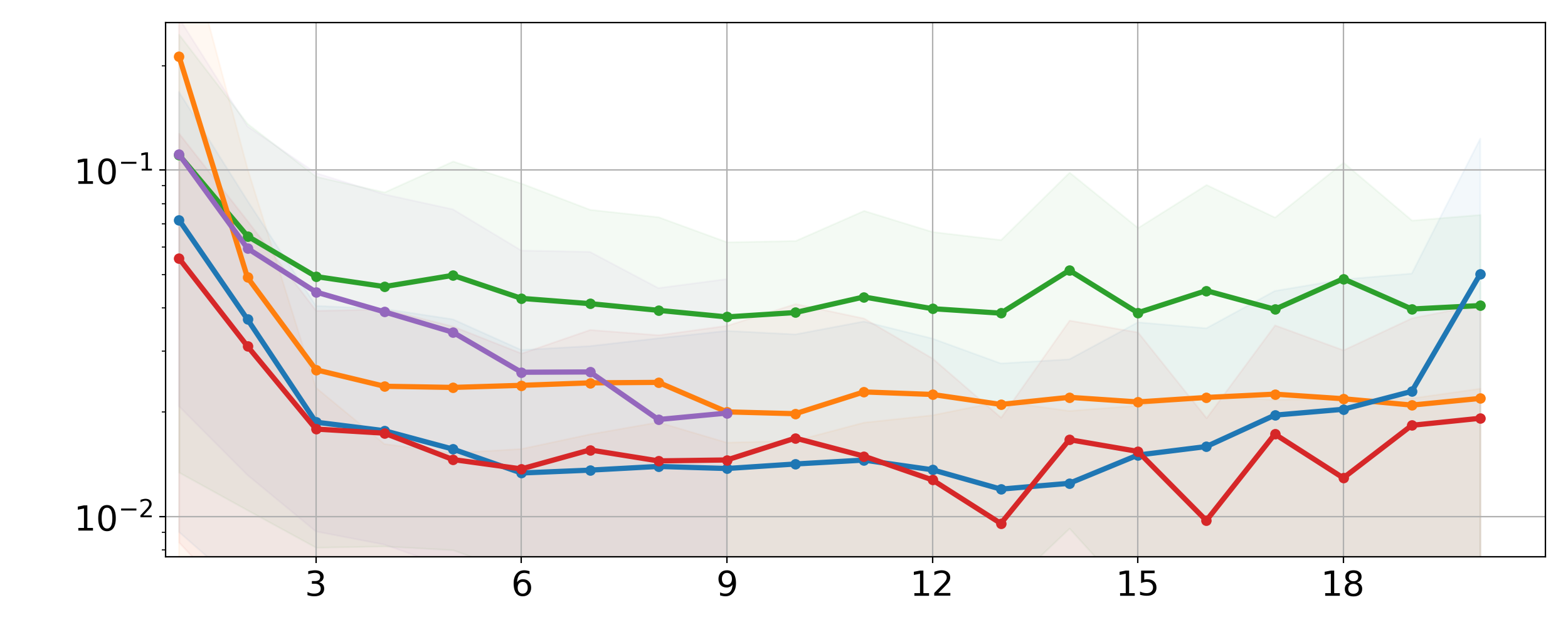}
\end{minipage}
\hfill
\begin{minipage}[t]{0.33\textwidth}
    \centering
    \includegraphics[width=\linewidth]{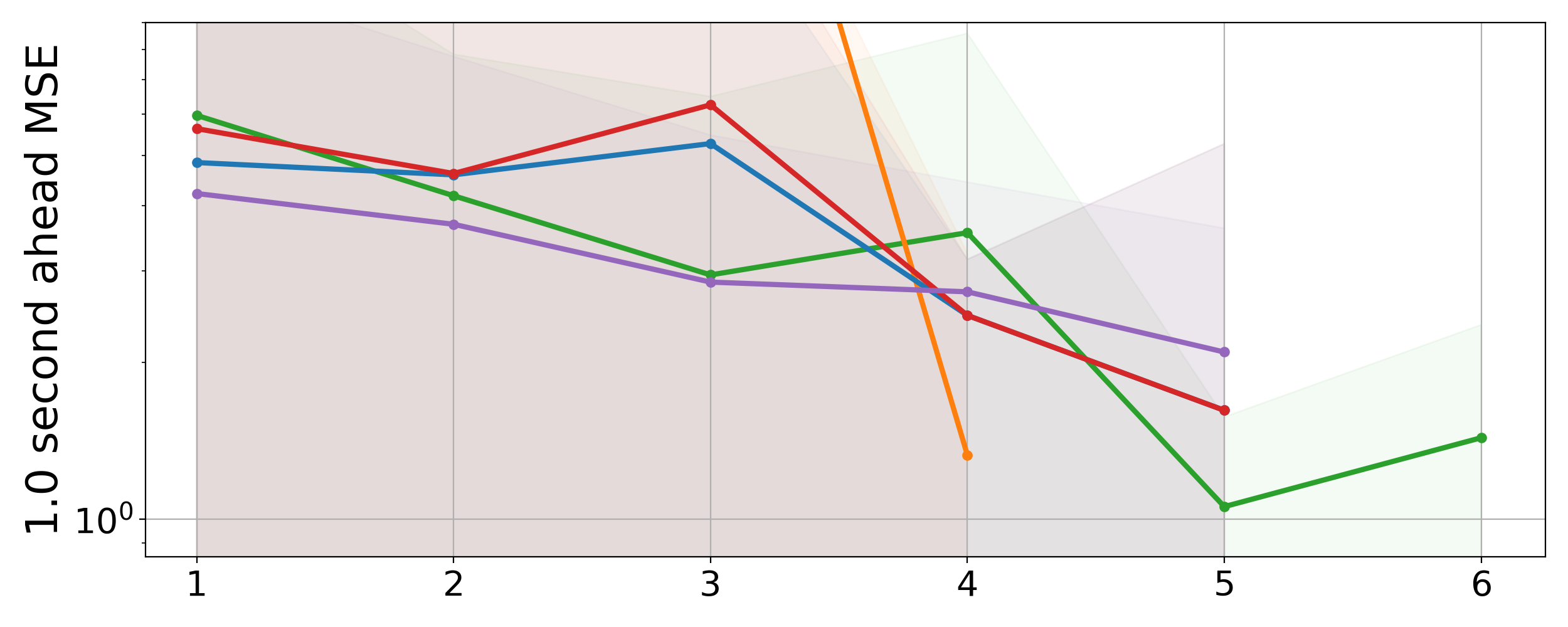}
\end{minipage}
\begin{minipage}[t]{0.33\textwidth}
    \centering
    \includegraphics[width=\linewidth]{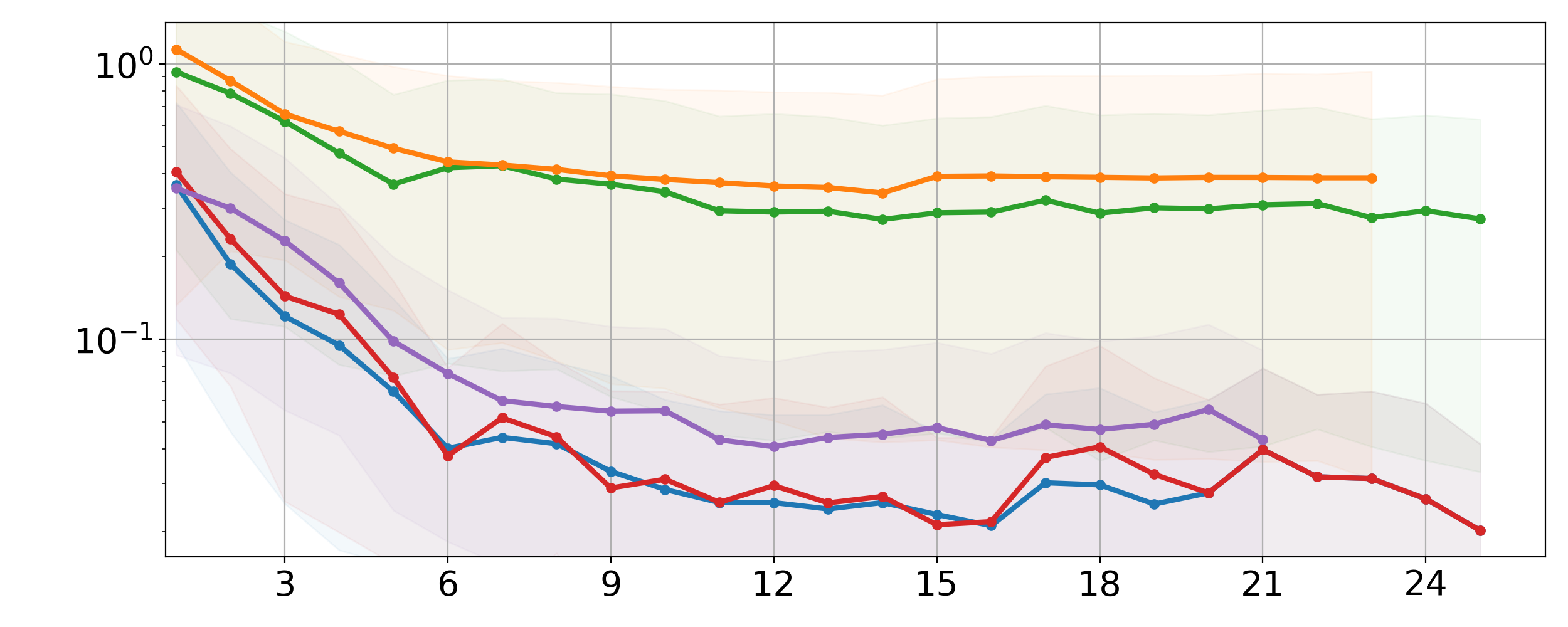}
\end{minipage}
\begin{minipage}[t]{0.33\textwidth}
    \centering
    \includegraphics[width=\linewidth]{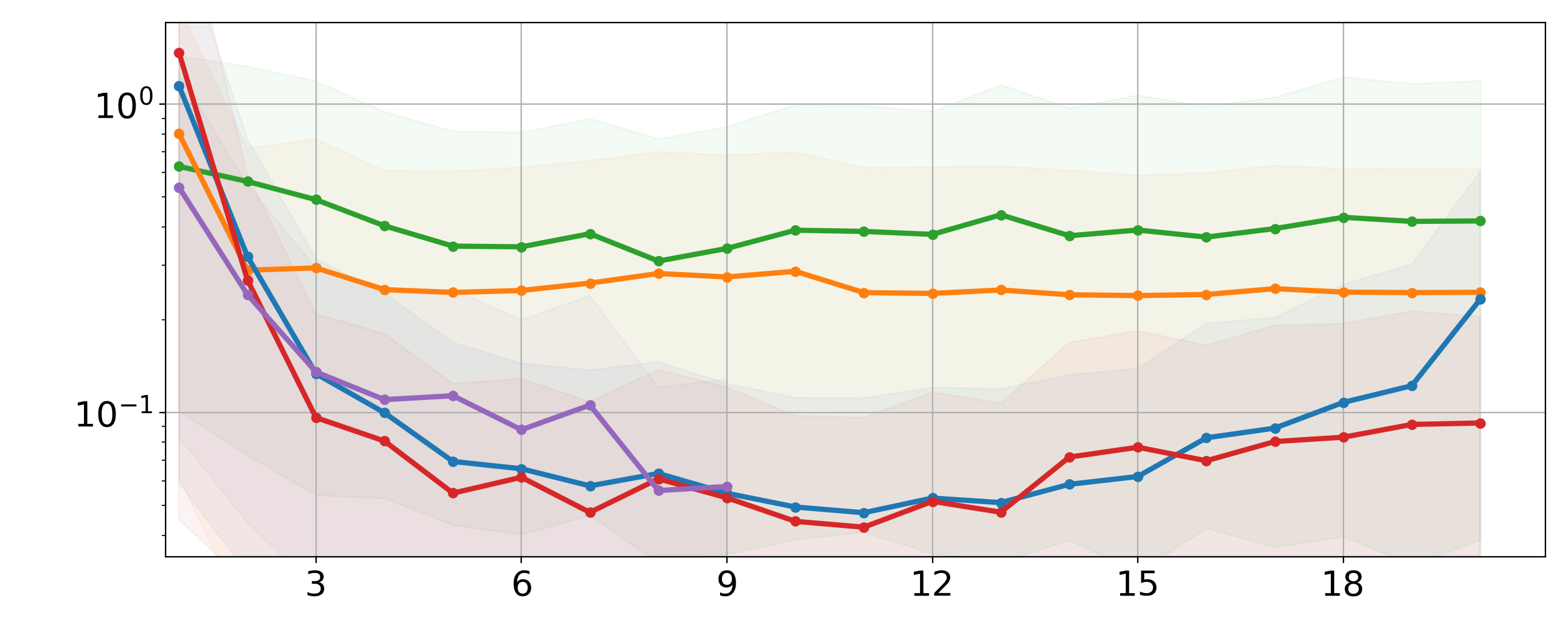}
\end{minipage}
\vspace*{-.5cm}
\caption{Predictive mean squared error of different dynamics models, computed after each round on a fixed test set.}
\label{fig:dyn}
\end{figure*}

Each ODE solver takes as input the same set of initial values as well as twenty different policy functions, some of which solve the problem whereas some are sub-optimal. Figure \ref{fig:ode-solver-comp} demonstrates the distance between the true state solutions and those given by different ODE solvers. The most striking observation is that discrete transitions of the form $\s_{t+1}-\s_t=h\cdot\f(\s_t,\a_t)$ are highly erroneous. Moreover, adaptive solvers as well as fixed-step solvers with sufficiently many intermediate steps attain practically zero error. Unsurprisingly, approximate state solutions deteriorate over time since the error accumulates. In our experiments, we use \texttt{RK78} to mimic the interactions with the real world, and \texttt{dopri5} to forward simulate model dynamics.

\section{Additional Related Work}

\paragraph{Model-based RL} The majority of model-based reinforcement learning methods assumes auto-regressive transitions, which effectively learn a distribution over the \textit{next} state given the current state and action. Unknown transitions are typically approximated by a Gaussian process \citep{kocijan2004gaussian, deisenroth2011pilco, kamthe2017data, levine2011nonlinear}, multi-layer perceptron (MLP) \citep{gal2016improving,depeweg2016learning,nagabandi2018neural,chua2018deep} or recurrent neural network \citep{ha2018world,hafner2018learning,hafner2019dream}. Such models are typically developed in conjunction with model predictive control \citep{richards2005robust} used for planning or with a parametric policy. 

\paragraph{Continuous-time RL} 
In addition to the works discussed earlier \citep{baird1993advantage,doya2000reinforcement}, \citet{bradtke1994reinforcement} developed Q-functions and temporal different learning in the context of semi-Markov decision processes. 
\citet{abu2005nearly} proposed a policy-iteration algorithm for the optimal control of CT systems with constrained controllers. An online version of this algorithm was derived in \cite{vrabie2009neural}, which was extended in a series of papers \citep{luo2014data,modares2016optimal,zhu2016using,lee2017integral}. 
A direct least-squares solution to Hamilton-Jacobi-Bellman equation was studied in \cite{tassa2007least}, requiring no forward integration for value estimation but a bag of tricks to deal with numerical instabilities. 
In a related work, \citet{mehta2009q} proposed an adaptive controller for nonlinear CT systems via a continuous-time analog of the Q-function. 
Above-mentioned methods are either built upon known dynamics or they are model-free. In either case, the dynamics are assumed to be linear with respect to the action, a premise needed for closed-form optimal policies.

\paragraph{Neural ODEs} In their ground-breaking work, \citet{chen2018neural} show that simple multi-layer perceptrons (MLP) can be utilized for learning arbitrary continuous-time dynamics. The resulting model, called Neural ODEs (NODEs), have been shown to outperform its RNN-based, discrete counterparts in interpolation and long-term prediction tasks \citep{chen2018neural}. The vanilla NODE model paved the way for advances in continuous-time modeling, such as second-order latent ODE models \citep{yildiz2019ode2vae}, augmented systems \citep{dupont2019augmented}, stochastic differential equations \citep{NEURIPS2019_59b1deff,pmlr-v108-li20i}, and so forth. NODE framework also allows encoding prior knowledge about the observed phenomena on the network topology, which leads to Hamiltonian and Lagrangian neural networks that are capable of long-term extrapolations, even when trained from images \citep{greydanus2019hamiltonian,cranmer2020lagrangian}.

\paragraph{Neural CTRL} The NODE breakthrough has opened a new research avenue in CTRL. In particular, physics-informed continuous-time dynamical systems have gained popularity. For example, Lagrangian mechanics are imposed on the architecture presented in \cite{lutter2019deep}, which results in near-perfect real-time control of a robot with seven degrees of freedom. Hamiltonian framework is proven useful for inferring controls from generalized coordinates and momenta \citep{zhong2019symplectic}. Later in \citet{zhong2020unsupervised}, an interpretable latent Lagrangian dynamical system and controller were trained from images. In addition to dynamics learning, above-mentioned methods describe model-specific recipes for learning controls. 




\newpage
\bibliographystyle{plainnat}
\bibliography{references}  

\end{document}